%% file: VLN.tex
\documentclass[final]{cvpr}

\usepackage{times}
\usepackage{epsfig}
\usepackage{graphicx}
\usepackage{amsmath}
\usepackage{amssymb}
\usepackage{multirow}
\usepackage{pifont}
\usepackage{booktabs}
\usepackage{epsfig}
\usepackage{booktabs,multirow}
\usepackage{graphics}
\usepackage{threeparttable}
\usepackage{color}
\usepackage[normalem]{ulem}
\usepackage{multirow}
\usepackage{float}
\usepackage{amsfonts}
\usepackage{bm}
\usepackage{wrapfig}
\usepackage{enumitem}
\usepackage[numbers]{natbib}
\usepackage{dblfloatfix}
\usepackage{array}
\usepackage[table]{xcolor}
\usepackage{colortbl}
\definecolor{myy}{RGB}{126,95,0}
\definecolor{mygray}{gray}{.9}
\definecolor{bblue}{RGB}{30,80,120}
\definecolor{mygray1}{gray}{.7}
\usepackage{bm}
\usepackage{mathtools}
\usepackage{subcaption}
\usepackage{siunitx}
\usepackage[nopar]{lipsum}
\usepackage[export]{adjustbox}
\usepackage{algorithm}
\usepackage{algpseudocode}

\newcolumntype{I}{!{\vrule width 1pt}}

\definecolor{ggray}{RGB}{127,127,127}
\definecolor{mygreen}{RGB}{93,174,86}
\definecolor{myred}{RGB}{192,0,0}
\definecolor{dgreen}{RGB}{55,86,35}

\makeatletter
\newcommand{\thickhline}{%
	\noalign {\ifnum 0=`}\fi \hrule height 1pt
	\futurelet \reserved@a \@xhline
}
\newcommand{\makesupptitle}[1]{
	\twocolumn[
	\begin{center}
		{\Large \bf #1 \par}
		\iftoggle{cvprrebuttal}{\vspace*{-22pt}}{\vspace*{24pt}}
		{
		\large
		\lineskip .5em
		\par
		}
		\vskip .5em
		\vspace*{12pt}
	\end{center}
	]
}
\makeatother
\newcommand{\tabincell}[2]{\begin{tabular}{@{}#1@{}}#2\end{tabular}}

\usepackage{caption}
\usepackage[pagebackref=true,breaklinks=true,colorlinks,bookmarks=false]{hyperref}
\usepackage[utf8]{inputenc}

\usepackage{cleveref}
\usepackage{subfiles}
\crefname{section}{§}{§§}
\Crefname{section}{§}{§§}

\def\cvprPaperID{2430} 
\def\confYear{CVPR 2022}
\begin{document}


\title{Counterfactual Cycle-Consistent Learning for Instruction Following and Generation in Vision-Language Navigation}

\author{Hanqing Wang$^{1,2}$~,~~\hspace{1pt}Wei Liang$^{1}$,~~\hspace{1pt}Jianbing Shen$^{3}$,~~Luc Van Gool$^{2}$,~~\hspace{1pt}Wenguan Wang$^{4}$\thanks{Corresponding author: \textit{Wenguan Wang}.}~   \\
	\small {$^1$} Beijing Institute of Technology
	{$^2$}  ETH Zurich
	{$^3$}  SKL-IOTSC, University of Macau
    {$^4$}  ReLER, AAII, University of Technology Sydney \hspace{0pt}\\
	\small \url{https://github.com/HanqingWangAI/CCC-VLN}
	%
}

\maketitle

\begin{abstract}
\vspace{-3pt}
Since$_{\!}$ the$_{\!}$ rise$_{\!}$ of$_{\!}$ vision-language$_{\!}$ navigation$_{\!}$ (VLN),$_{\!}$ great progress has been made in \textbf{instruction following} -- building$_{\!}$ a follower$_{\!}$ to$_{\!}$ navigate$_{\!}$ environments$_{\!}$ under the guidance$_{\!}$ of$_{\!}$ instructions.$_{\!}$ However,$_{\!}$ far$_{\!}$ less$_{\!}$ attention$_{\!}$ has$_{\!}$ been$_{\!}$ paid$_{\!}$ to$_{\!}$ the inverse task: \textbf{instruction generation} -- learning a \textit{speaker}~to generate grounded descriptions for navigation routes.~Exi- sting VLN methods train a speaker independently and often treat it as a data augmentation tool to strengthen the fol- lower,$_{\!}$ while$_{\!}$ ignoring$_{\!}$ rich$_{\!}$ cross-task$_{\!}$ relations.$_{\!}$ Here$_{\!}$ we$_{\!}$~des-  cribe an approach that learns the two tasks simultaneously and exploits their intrinsic correlations to boost the training of each: the follower judges whether the speaker-created instruction explains the original navigation route correctly, and vice versa.$_{\!}$  Without the need of aligned instruction-path pairs, such cycle-consistent learning scheme is complementary to task-specific training targets defined on labeled data, and can also be applied over unlabeled paths$_{\!}$ (sampled without paired instructions). Another agent, called~creator   is added to generate counterfactual environments. It greatly changes current scenes yet leaves novel items -- which are vital for the execution of original instructions -- unchanged. Thus more informative training scenes are synthesized and the three agents compose a powerful VLN learning system. Extensive experiments on a standard benchmark
show that our approach improves the performance of various follower models and produces accurate navigation instructions.
\end{abstract}

\vspace{-9pt}

\section{Introduction}\label{sec:1}

\begin{figure}[t]
  \centering
  \vspace{-2pt}
      \includegraphics[width=0.99\linewidth]{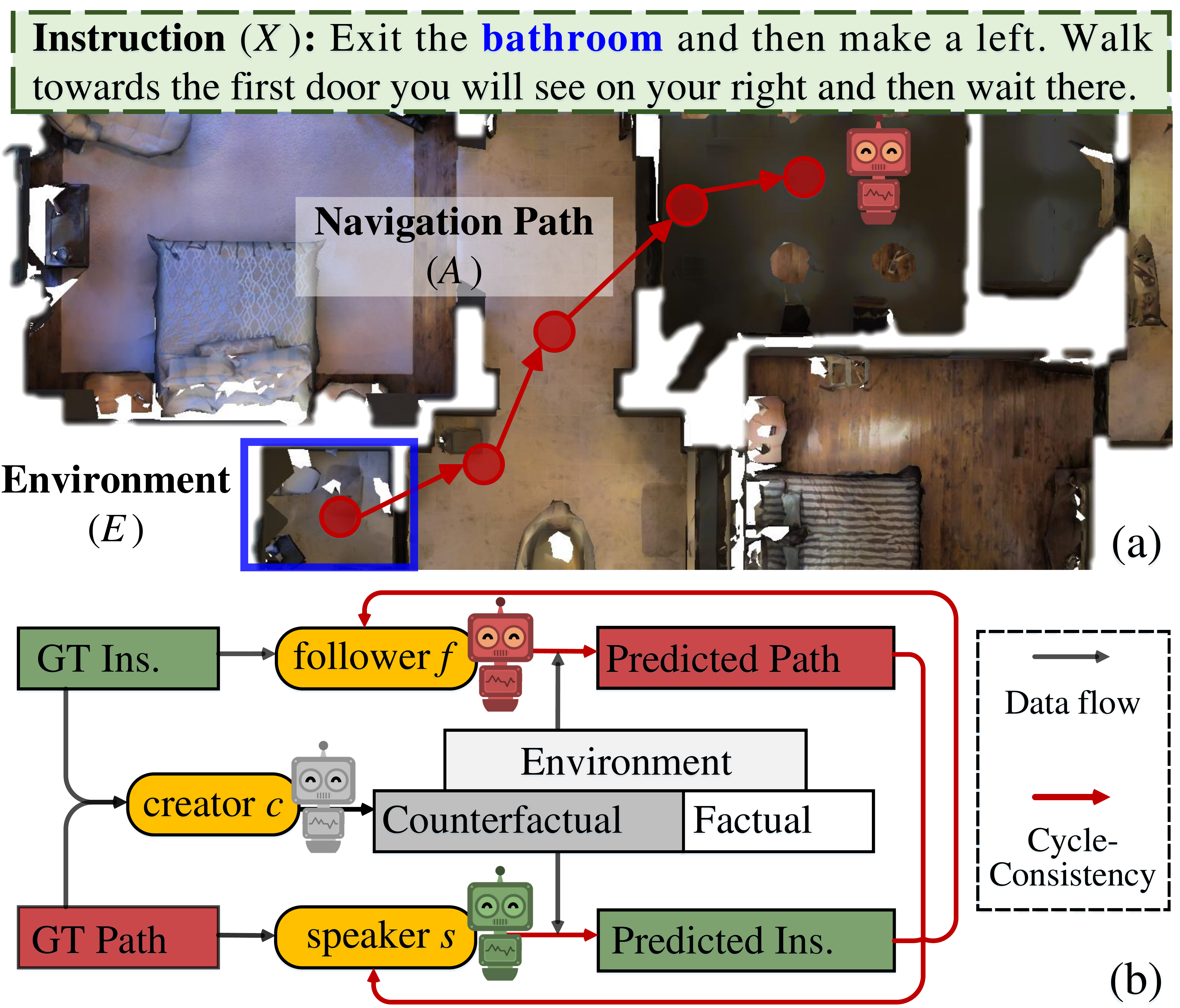}
    \put(-200.5,66.5){{\footnotesize$X$}}
    \put(-67,67){{\footnotesize$\hat{A}$}}
    \put(-110,35){{\footnotesize$E$}}
    \put(-72.5,35){{\footnotesize$\bar{E}$}}
    \put(-199.5,11.5){{\footnotesize$A$}}
    \put(-68.5,11){{\footnotesize$\hat{X}$}}
\vspace{-7pt}
\captionsetup{font=small}
	\caption{\small{(a) VLN~\cite{anderson2018vision}. (b) Our counterfactual cycle-consistent (CCC) learning system consists of three agents, \ie, speaker $s$ (\protect\includegraphics[scale=0.07,valign=c]{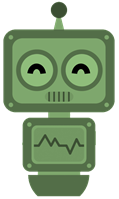}) for instruction generation, follower $f$ (\protect\includegraphics[scale=0.07,valign=c]{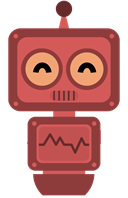}) for instruction following, and creator $c$  (\protect\includegraphics[scale=0.07,valign=c]{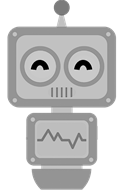}) for counterfactual environment creation.  
}
}
\label{fig:overview}
\vspace{-15pt}
\end{figure}

Vision-language navigation (VLN)~\cite{anderson2018vision}, \ie, enabling an agent to navigate across realistic environments given human instructions,  has received great attention (Fig.~\ref{fig:overview}(a)). Many powerful wayfinding agents (\ie, \textit{follower}) were developed to perform such embodied \textbf{instruction following} task. Unfortunately, the inverse task -- learning$_{\!}$ a$_{\!}$ \textit{speaker}$_{\!}$ that vividly explains navigation paths -- has remained under-explored.$_{\!}$ In$_{\!}$ fact, \textbf{instruction generation} is also a crucial ability of AI agents. In many scenarios, AI agents should be able to communicate with$_{\!}$ humans for$_{\!}$ efficient$_{\!}$ collaboration,$_{\!}$ instead$_{\!}$ of$_{\!}$ executing$_{\!}$ instructions$_{\!}$ only~\cite{wang2007human,green2008human}. For example, when a$_{\!}$ human-robot$_{\!}$ team$_{\!}$ is$_{\!}$ doing$_{\!}$ search$_{\!}$ and$_{\!}$ rescue$_{\!}$~\cite{tellex2011understanding,daniele2017navigational,wang2021collaborative}, the  human may first issue commands$_{\!}$ (\eg,$_{\!}$ ``explore$_{\!}$ along$_{\!}$ this direction until the end of the hallway'') that direct the robot to$_{\!}$ navigate a building and search for$_{\!}$ survivors.$_{\!}$ In$_{\!}$ this$_{\!}$ process,$_{\!}$ the robot is expected to report its progress (\eg, ``I have inspected three rooms'') and explain its plan (\eg, ``I will continue to navigate this direction and stop at the end of the hallway'')$_{\!}$~\cite{thomason2020vision}. Robot's ability to generate linguistic explanations can help human to resolve potential ambiguity (\eg, identifying ``this direction'' and ``hallway'')$_{\!}$~\cite{daniele2017navigational} and establish trust~\cite{andrist2013rhetorical,dzindolet2003role}. Hence,$_{\!}$ the$_{\!}$ robot$_{\!}$ can$_{\!}$ even$_{\!}$ in$_{\!}$ turn$_{\!}$ help the human to navigate its explored area (which is unfamiliar to the human)$_{\!}$~\cite{goeddel2012dart}, \eg, ``go straight and you will pass through four rooms before reaching the end of the hallway''.

In addition to emphasizing the importance of instruction generation, we$_{\!}$ explore$_{\!}$ the$_{\!}$ intrinsic$_{\!}$ correlation between instruction following and generation to derive a powerful VLN learning framework. Specifically, given the visual environment space $\mathcal{E}$, linguistic instruction space $\mathcal{X}$, and navigation path space $\mathcal{A}$, instruction following learns a follower $f\!:\! \mathcal{E}\!\times\!\mathcal{X}\!\mapsto\!\mathcal{A}$ that maps visual observations and navigable directions into action sequences, while instruction generation learns a speaker $s\!:\! \mathcal{E}\!\times\!\mathcal{A}\!\mapsto\!\mathcal{X}$ that maps observations and$_{\!}$ action sequences into trustable instructions. Clearly, there exists strong dependencies among the input and output spaces of $s$ and $f$.  Surprisingly, such task correlation was long ignored; current VLN methods only learn an isolated speaker as an one-off plugin for data augmentation~\cite{fried2018speaker,tan2019learning}.

We instead propose to jointly train the speaker and follower in a compact, cycle-consistent learning framework  (Fig.\!~\ref{fig:overview}(b)). During training, $(E, X)\!\in\!\mathcal{E}\!\times\!\mathcal{X}$ is first mapped to $\hat{A}\!\in\!\mathcal{A}$ through the follower $f$ (\includegraphics[scale=0.07,valign=c]{figure/follower}) and then translated to an instruction $\tilde{X}\!\in\!\mathcal{X}$ through the speaker (\includegraphics[scale=0.07,valign=c]{figure/speaker}), \ie, $s(E, \hat{A})$. In environment $E$, the dissimilarity between $X$ and $\tilde{X}$, denoted as $\triangle_E(X,\tilde{X})$, is used as the feedback signal to regularize training. Similarly, given $(E, A)\!\in\!\mathcal{E}\!\times\!\mathcal{A}$, $\triangle_E(A, \tilde{A})$ can be estimated and used for training. As $\triangle$ errors are only about the cycle-consistency over $(E, X)$ and $(E, A)$, any other training objectives defined on labeled triplets, \ie, $(E, X, A)$, are compatible. Hence, we can apply such learning system on ``unlabeled'' data $(E, A')$, \ie, sampling a path $A'$ in an environment $E$ without corresponding instruction.  Thus both labeled instruction-path samples and unlabeled paths can be simultaneously used during training. This is more elegant than current \textit{de facto} VLN training protocol~\cite{fried2018speaker,tan2019learning} that has three phases: i) train the follower and speaker separately on
aligned instruction-path samples; ii) use the speaker to create synthetic instructions for randomly sampled paths; and iii) fine-tune the follower on the pseudo instruction-path samples. Further, as  learning from pseudo-parallel data inevitably accompanies with
the data quality problem (\ie, the quality of the pseudo instruction-path samples is difficult to guarantee)~\cite{luo2019dual},  our speaker-follower collaborative learning game is more favored, \ie, $\triangle$ errors can be viewed as quality scores and play as supervisory signals to boost the training of both $f$ and $s$.

Besides$_{\!}$ the$_{\!}$ speaker$_{\!}$ and$_{\!}$ follower, another agent, called \textit{creator} (\includegraphics[scale=0.07,valign=c]{figure/creator}),$_{\!}$ is put$_{\!}$ into$_{\!}$ our$_{\!}$ cycle-consistent$_{\!}$ learning$_{\!}$ game. The$_{\!}$ creator$_{\!}$ serves$_{\!}$ as$_{\!}$ a$_{\!}$ plug-and-play$_{\!}$ component$_{\!}$ for$_{\!}$ \textbf{counterfactual environment synthesis}, enabling more robust training.$_{\!}$ Thinking$_{\!}$ about$_{\!}$ alternative$_{\!}$ possibilities$_{\!}$ for past or future events is central to human thinking~\cite{roese1997counterfactual}. We frequently construct counterfactuals (``counter to the facts''): what might happen if $\cdots$? Counterfactual thinking gives us the flexibility in learning from past limited experience through mental simulation. Nevertheless, this issue is rarely addressed in VLN. Recently, \cite{fu2019counterfactual} interpreted current prevailing data augmentation$_{\!}$ technique$_{\!}$~\cite{fried2018speaker} -- back-translation -- as$_{\!}$ a$_{\!}$ kind$_{\!}$ of$_{\!}$ counterfactual$_{\!}$ thoughts -- given$_{\!}$ an$_{\!}$ instruction$_{\!}$ like$_{\!\!}$ ``walk$_{\!}$ away$_{\!}$ from$_{\!}$ the$_{\!}$ door'',$_{\!}$ the$_{\!}$ agent$_{\!}$ builds$_{\!}$ a$_{\!}$ counterfa- ctual like ``if I walk in the door, what should the instruction be?'' by augmenting sampled paths with artificial instructions. This allows the agent to be more efficiently trained by performing alternative actions that it did not actually make$_{\!}$~\cite{fu2019counterfactual}. Although our speaker-follower game naturally supports such path sampling based counterfactual thinking (\ie, involving unlabeled data $(E, A')$ during$_{\!}$ training,$_{\!}$ as$_{\!}$ mentioned$_{\!}$ before), our creator$_{\!}$ can build another kind of counterfactuals, \eg, ``if I change current environment by, for example, removing or putting in a lot of furniture that are irrelevant to the instruction, I will still execute the original navigation plan''. Concretely, given an environment-instruction-path tuple $(E, X, A)_{\!}\in_{\!}\mathcal{E}_{\!}\times_{\!}\mathcal{X}_{\!}\times_{\!}\mathcal{A}$, the creator $c{\!}:{\!} \mathcal{E}{\!}\times{\!}\mathcal{X}{\!}\times{\!}\mathcal{A}{\!}\mapsto{\!}\mathcal{E}$ ``images'' a new environment $\bar{E}_{\!}\in_{\!}\mathcal{E}$, such that i) $\bar{E}$ is greatly different from $E$, and ii) $\bar{E}$ should be still aligned with the original instruction-path pair $(X, A)$. With i) and ii), we address both diversity and realism, which are proved critical in counterfactual thinking~\cite{girotto2007postdecisional,woodward2011psychological}. Hence, as $\bar{E}$ and $(X, A)$ compose a new valid training sample, the cycle-consistent errors, \ie,  $\triangle_{\bar{E}}(A,\tilde{A})$ and $\triangle_{\bar{E}}(X,\tilde{X})$, can be estimated over $(\bar{E}, X, A)$. In this way, our speaker, follower and creator form a powerful learning system, which makes a clever use of cross-task and cross-modal connections as well as resembles a counterfactual thinking process.

Our counterfactual cycle-consistent (CCC) framework is optimized by policy gradient methods and compatible with current imitation learning (IL) and reinforcement learning (RL)$_{\!}$ based$_{\!}$ VLN training$_{\!}$ protocols.$_{\!}$ We$_{\!}$ apply$_{\!}$ our CCC over several VLN baseline models and test it on gold-standard R2R$_{\!}$ dataset$_{\!}$~\cite{anderson2018vision}.$_{\!}$ Experimental results verify the efficacy of CCC on both instruction following$_{\!}$ and$_{\!}$ generation$_{\!}$ tasks.

\section{Related Work}%
\noindent\textbf{Vision-Language Navigation (Instruction Following).} Although VLN$_{\!}$~\cite{anderson2018vision} is a relatively new task in computer vision, its core part -- instruction following$_{\!}$~\cite{chen2011learning,mei2016listen} -- has been long studied in natural language processing and robotics. Early studies typically built the navigator in controlled environments with formulaic route descriptions$_{\!}$~\cite{macmahon2006walk,tellex2011understanding,chen2011learning,andreas2015alignment,mei2016listen,misra2018mapping}. Anderson \etal thus introduced R2R dataset$_{\!}$~\cite{anderson2018vision} to investigate embodied navigation in photo-realistic simulated environments$_{\!}$~\cite{Matterport3D} with human-created instructions. Soon after, numerous efforts were made towards: \textbf{i)}
raising more efficient learning paradigms, \eg, IL~\cite{anderson2018vision}, hybrid of model-free and model-based RL$_{\!}$~\cite{wang2018look}, and ensemble of IL and RL$_{\!}$~\cite{wang2019reinforced}; \textbf{ii)} exploring extra supervisory signals from synthesized samples$_{\!}$~\cite{fried2018speaker,tan2019learning,fu2019counterfactual}, auxiliary tasks$_{\!}$~\cite{wang2019reinforced,huang2019transferable,ma2019self,zhu2019vision}, or even massive web image-text paired data$_{\!}$~\cite{majumdar2020improving,hao2020prevalent}; \textbf{iii)}  developing more powerful perception-linguistic embedding schemes$_{\!}$~\cite{hu2019you,qi2020object,wang2020environment,Hong_2021_CVPR}; and \textbf{iv)} designing smarter path planning strategies by self-correction$_{\!}$~\cite{ke2019tactical,ma2019regretful}, active exploration$_{\!}$~\cite{wang2020active}, or map building$_{\!}$~\cite{wang2021structured,Chen_2021_CVPR,deng2020evolving}. Some others learn environment-agnostic representations$_{\!}$~\cite{wang2020environment}, or focus on fine-grained instruction parsing$_{\!}$~\cite{hong2020language}.

Our work differs significantly from the above body of work. We address the importance of both instruction following and generation, instead of the navigation task only.  The dual tasks form a closed loop for end-to-end joint training. We are particularly interested in how to leverage their intrinsic connections to better learn each other, within factual environments as well as counterfactual alternatives.

\noindent\textbf{Instruction Generation.}$_{\!}$  Though$_{\!}$ being$_{\!}$ less$_{\!}$ studied$_{\!}$ in computer vision~\cite{agarwal2019visual}, generating linguistic route instructions$_{\!}$~\cite{curry2015generating}$_{\!}$ has$_{\!}$ raised$_{\!}$ wide$_{\!}$ research$_{\!}$ interest in robotics~\cite{goeddel2012dart}, linguistics\!~\cite{striegnitz2011report}, cognition\!~\cite{kuipers1978modeling}, and environmental psychology\!~\cite{vanetti1988communicating}, and$_{\!}$ can$_{\!}$ be$_{\!}$ traced$_{\!}$ to$_{\!}$ Lynch's$_{\!}$ work$_{\!}$~\cite{Lynch}$_{\!}$ in$_{\!}$ 1960. Early~efforts investigated the principles underlying the process of human constructing route descriptions$_{\!}$~\cite{ward1986turn,allen1997knowledge,lovelace1999elements} and the characteristics  of ``easy-to-follow'' instructions$_{\!}$~\cite{look2005location,waller2007landmarks,richter2008simplest}. They pointed out the importance of involving intuitive landmarks (\eg, physical objects and locations) and concise topological descriptions (\eg, turn-by-turn directions) in instructions$_{\!}$~\cite{daniele2017navigational}. Based on these studies, some simple systems$_{\!}$~\cite{look2005location,goeddel2012dart} for instruction creation were developed using handcrafted
\textit{templates}, \ie, slotting the content into pre-built linguistic structures. Some complicated ones$_{\!}$~\cite{dale2004using} made use of linguistically motivated \textit{rules} or full-fledged \textit{grammars}, to better emulate the way people compose instructions and produce outputs in a more flexible and extensible manner$_{\!}$~\cite{foster2019natural}. Recent solutions$_{\!}$~\cite{cuayahuitl2010generating,osswald2014learning,daniele2017navigational,fried2017unified} lean on end-to-end, data-driven techniques,  without manually crafted templates or rules. But they are$_{\!}$ typically performed$_{\!}$ on$_{\!}$ simple grid or rendered environments and thus factor out the~role of perception in instruction creation to some extent.

Instruction creation attains far less attention in VLN~\cite{agarwal2019visual}.
Only some$_{\!}$ data augmentation$_{\!}$ techniques\!~\cite{fried2018speaker,tan2019learning,fu2019counterfactual} learn$_{\!}$ an instruction generator (speaker) with training pairs of aligned navigation paths and human instructions. Then they use the speaker to synthesize instructions for newly sampled paths as extra training examples. However, they only treat instruction creation as an auxiliary task and train the speaker and follower separately. Hence it is hard to control the quality of the pseudo instructions created by the speaker. Unlike these methods, we propose a unified framework that learns the speaker and follower simultaneously, and  explicitly uses their correlation as a regularization term for robust training.

\noindent\textbf{Cycle-Consistent Learning.} Cycle-consistent learning explores task correlations to regularize training and can be implemented in different forms, such as forward-backward object tracking~\cite{wang2019unsupervised,lu2020learning}, CycleGAN~\cite{zhu2017unpaired}, and dual learning~\cite{he2016dual}. Taking dual learning as an example, its idea is intuitive: if we map an $x$ from one domain to another and then map it back, we should recover
the original $x$~\cite{zhao2020dual}.  It was successfully applied to many tasks like neural machine translation~\cite{he2016dual}, sentiment analysis~\cite{xia2017dual}, image-to-image translation~\cite{kim2017learning}, question answering~\cite{tang2017question,shah2019cycle,li2018visual}, \etc.

In a broader sense, our study can be viewed as the first attempt that explores the duality of instruction generation and following in embodied navigation tasks. Both the two tasks are learnt in a dual-task learning framework, where their symmetric structures are explored as informative feedback signals for boosting each, even with unlabeled samples.

\noindent\textbf{Counterfactual Thinking.} Counterfactual thinking~\cite{roese1997counterfactual} (\ie, the construction of mental alternatives to reality) is crucial to how people learn from experience and predict the future, and can influence different cognitive behaviors such as deduction, decision making, and problem solving~\cite{yang2021multiple,epstude2008functional}. Recent studies proved that counterfactual thoughts can improve the explainability~\cite{hendricks2018grounding,goyal2019counterfactual}, fairness~\cite{kusner2017counterfactual}, and robustness~\cite{wang2020visual} of trained models. The use of counterfactual examples has also been explored in the context of visual question answering~\cite{agarwal2020towards,abbasnejad2020counterfactual,chen2020counterfactual} and open set recognition~\cite{neal2018open,yue2021counterfactual}.

Fu \etal~\cite{fu2019counterfactual} revisit the$_{\!}$ idea$_{\!}$ of$_{\!}$ back-translation$_{\!}$~based data$_{\!}$ augmentation from a counterfactual thinking
perspective: extra routes are adversarially selected, instead of randomly sampled~\cite{fried2018speaker}, and translated into instructions for smarter$_{\!}$ data$_{\!}$ augmentation.$_{\!}$ Apart$_{\!}$ from$_{\!}$ sampling$_{\!}$ paths$_{\!}$ in$_{\!}$ real$_{\!}$ environments,$_{\!}$ we$_{\!}$ learn$_{\!}$ a$_{\!}$ creator$_{\!}$ to$_{\!}$ generate new visual scenes as more effective counterfactuals. Although counterfactual environment synthesis is also addressed in~\cite{parvaneh2020counterfactual}, it is achieved by introducing$_{\!}$ minimum$_{\!}$ interventions$_{\!}$ that$_{\!}$ cause$_{\!}$ the$_{\!}$ follower$_{\!}$ to$_{\!}$ change$_{\!}$ its$_{\!}$ output.$_{\!}$ In$_{\!}$ contrast,$_{\!}$ we$_{\!}$~seek$_{\!}$ to$_{\!}$~mo- dify$_{\!}$ the$_{\!}$ factual$_{\!}$ environments$_{\!}$ to$_{\!}$ maximum, on$_{\!}$ the$_{\!}$ premise$_{\!}$ of$_{\!}$ ensuring$_{\!}$ the$_{\!}$ original$_{\!}$ instructions$_{\!}$ can$_{\!}$ be$_{\!}$ still$_{\!}$ executed. The$_{\!}$ combination$_{\!}$ of$_{\!}$ diversity$_{\!}$ and$_{\!}$ realism$_{\!}$ makes$_{\!}$ generated$_{\!}$ environments$_{\!}$ useful$_{\!}$ as$_{\!}$ training$_{\!}$ examples. Moreover,$_{\!}$ we$_{\!}$ learn$_{\!}$ the speaker$_{\!}$ and follower end-to-end collaboratively.\!

\begin{figure*}[!tbh]
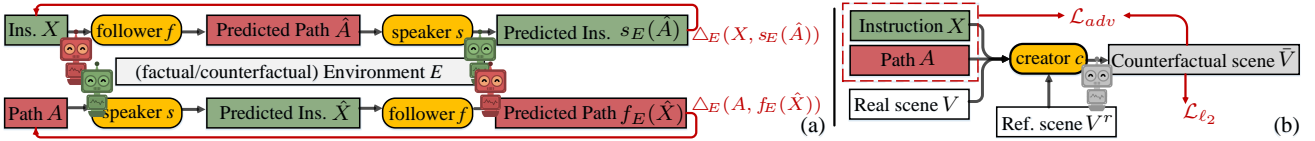

  \centering
\vspace{-6pt}
      \includegraphics[width=0.99\linewidth]{figure/fig2}
      \put(-477,40){{\footnotesize$X$}}
      \put(-475,8){{\footnotesize$A$}}
      \put(-366,39.5){{\footnotesize$\hat{A}$}}
      \put(-367,8){{\footnotesize$\hat{X}$}}
      \put(-258,40){{\footnotesize$s_E(\hat{A})$}}
      \put(-258,8){{\footnotesize$f_E(\hat{X})$}}
      \put(-231,38.5){\textbf{\color{myred}\scriptsize$\triangle_{\!E}(X, s_{\!E}(\hat{A}))$}}
      \put(-231,12){\textbf{\color{myred}\scriptsize$\triangle_{\!E}(A, f_{\!E}(\hat{X}))$}}
      \put(-134.5,41){{\footnotesize$X$}}
      \put(-145,28.8){{\footnotesize$A$}}
      \put(-136.2,11.6){{\footnotesize$V$}}
      \put(-9,28.5){{\footnotesize$\bar{V}$}}
      \put(-88,45){\color{myred}{\footnotesize$\mathcal{L}_{adv}$}}
      \put(-45,9){\color{myred}{\footnotesize$\mathcal{L}_{\ell_2}$}}
      \put(-82,4.5){{\footnotesize$V^r$}}
\vspace{-6pt}
\captionsetup{font=small}
	\caption{\small{(a) Our cycle-consistent learning scheme (\textit{cf.}~\S\ref{sec:2ag}) for the speaker $s$ (\protect\includegraphics[scale=0.07,valign=c]{figure/speaker}) and follower $f$ (\protect\includegraphics[scale=0.07,valign=c]{figure/follower}), trained over both factual and counterfactual environments. (b) Our creator  (\protect\includegraphics[scale=0.07,valign=c]{figure/creator}) for counterfactual environment generation (\textit{cf.}~\S\ref{sec:CEC}).
}
}
\label{fig:method}
\vspace{-12pt}
\end{figure*}

\vspace{-3pt}
\section{Methodology}
\vspace{-2pt}
We address two related tasks, \ie, instruction following and generation,$_{\!}$ under$_{\!}$ R2R$_{\!}$ VLN$_{\!}$ setting$_{\!}$~\cite{anderson2018vision}.$_{\!}$ For$_{\!}$ instruction following, a follower $f$ is learnt to find a route $A$ to the target location, specified by the instruction $X$, in a 3D environment $E$. For instruction generation, a speaker $s$  is learnt to$_{\!}$ create$_{\!}$ a$_{\!}$ description$_{\!}$ $X$$_{\!}$ for$_{\!}$ route$_{\!}$ $A$$_{\!}$ in$_{\!}$ $E$.$_{\!}$ Here,$_{\!}$ $s$$_{\!}$ and$_{\!}$ $f$$_{\!}$ are jointly trained in an end-to-end cycle-consistent framework (\textit{cf.}~\S\ref{sec:2ag}) and a creator $c$ is further introduced to synthesize counterfactual environments for boosting training (\textit{cf.}~\S\ref{sec:CEC}).

\vspace{-1pt}
\subsection{Cycle-Consistent Learning for Instruction Following and
Generation}\label{sec:2ag}\vspace{-1pt}
We learn the two tasks jointly (Fig.~\ref{fig:method}(a)): the speaker $s$ and follower $f$ act as an evaluator for each other. $s$ is used to evaluate the quality of $\hat{A}$ generated by $f(E, X)$ and returns the feedback signal $\triangle_E(X, s(E, \hat{A}))$ to $f$, and vice versa.

\noindent\textbf{Follower.} The follower $f$ is instantiated as a {\texttt{Seq2Seq}} model which computes a distribution $P(A|X; E)$ over route $A$ (\ie, a serious of actions $A\!=\!\{a_t\}_{t=1}^T$) given instruction $X$ (\ie, a sequence of words $X\!=\!\{x_l\}_{l=1}^L$) under environment $E$. At each step $t$, the follower observes $E$ as an image scene $V_t\!\subset\!E$.  Conditioned
on the visual and linguistic features, \ie,  $\bm{V}_t$ and $\bm{X}\!=\!\{\bm{x}_l\}_{l=1}^L$, and prior action embedding $\bm{a}_{t-1}$, the follower $f$ first produces current hidden state $\bm{h}^f_t$:\!
\vspace{-2pt}
\begin{equation}\small\label{eq:1}
\begin{aligned}
\bm{h}^f_t = \text{LSTM}^f([\bm{V}_{t-1}, \bm{X}, \bm{a}_{t-1}], \bm{h}^f_{t-1}).
\end{aligned}
\vspace{-2pt}
\end{equation}
There are two basic follower designs in the literature, based on their definitions of the action space. The first-type followers~\cite{anderson2018vision} simplify the action space to six
low-level \textit{visuomotor} behaviors, \ie, \texttt{left}, \texttt{right}, \texttt{up}, \texttt{down}, \texttt{forward} and \texttt{stop}. For example, \texttt{left} refers to turning left by $30^\circ$. The action embeddings are linguistic features and only a front view is perceived as $V_{t}$. Given proceeding actions $a_{1:t_{\!}-_{\!}1}$, instruction$_{\!}$ $X$$_{\!}$ and past$_{\!}$ observations$_{\!}$ $V_{1:t_{\!}-_{\!}1}$,$_{\!}$ the$_{\!}$ conditional probability of current action $a_t$ is computed as:
\vspace{-2pt}
\begin{equation}\small\label{eq:2}
\begin{aligned}
P(a_t|a_{1:t-1}, x_{1:L}, V_{1:t-1}) = \text{softmax}_{a_t}(\bm{W}_1\bm{h}^f_t),
\end{aligned}
\vspace{-2pt}
\end{equation}
where $\bm{W}_1\bm{h}_t\!\in\!\mathbb{R}^6$ gives a score vector over the six actions.

The second-type followers~\cite{fried2018speaker,wang2019reinforced} first ``look around'' and gain a \textit{panoramic} view as $V_{t}$. Then $V_{t}$ is divided into 36 subviews, \ie, $V_{t}\!=\!\{V_{t,i}\}_{i=1}^{36}$, forming current action space.  Thus each action $a_t$ is associated with a subview $V_{t, a_t}$, and $\bm{a}_t\!\equiv\!\bm{V}_{t, a_t}$. Then, the likelihood of $a_t$ is formulated as:
\vspace{-2pt}
\begin{equation}\small\label{eq:3}
\begin{aligned}
P(a_t|a_{1:t-1}, x_{1:L}, V_{1:t-1}) = \text{softmax}_{a_t}(\bm{a}_t^\top\bm{W}_2\bm{h}_t).
\end{aligned}
\vspace{-2pt}
\end{equation}
As$_{\!}$ the$_{\!}$ percepts$_{\!}$ $V_{1:t-1\!}$ are$_{\!}$ completely$_{\!}$ determined$_{\!}$ by$_{\!}$ the navigation actions $a_{1:t-1}$ in all our cases, we can further specify $P(A|X; E)$ according to the probability chain rule: $P(A|X; E)\!=\!\prod_{t\!}P(a_t|a_{1:t-1}, x_{1:L},_{\!} V_{1:t-1})$. And for notational simplicity, we abbreviate the probability distribution $P(a_t|a_{1:t-1}, x_{1:L}, V_{1:t-1})$ over actions at step $t$ as $p_t(a_t)$.

To fully examine the  efficacy of our CCC framework,  we experiment with different follower architectures~\cite{anderson2018vision,fried2018speaker,wang2019reinforced}.

\noindent\textbf{Speaker.}  The speaker $s$ is built as a recurrent neural network based encoder-decoder architecture which computes a distribution $P(X|A; E)$ over possible instruction $X$ ($=\!\{x_l\}_{l=1}^L$) given route $A$ in environment $E$.

The encoder first embeds the sequences of the actions  $\{a_t\}_{t=1}^T$ and visual observations $\{V_t\}_{t=1}^T$ along the route, and generates hidden states $\bm{o}^s_{1:T}$ using an LSTM:
\vspace{-2pt}
\begin{equation}\small\label{eq:4}
\begin{aligned}
\bm{o}^s_t = \text{LSTM-Encoder}^s([\bm{V}_{t}, {\bm{a}}_t], \bm{o}^s_{t-1}).
\end{aligned}
\vspace{-2pt}
\end{equation}
Afterwards, the decoder computes the conditional probability of each target word $x_l$ given its proceeding words $x_{1:l-1}$ as well as the input embedding $\bm{o}^s_{T}$:\!\!
\vspace{-2pt}
\begin{equation}\small\label{eq:5}
\begin{aligned}
\!\!P(x_l|x_{1:l-1}, a_{1:T}, V_{1:T}) \!= \!\text{LSTM-Decoder}^s(\bm{h}^s_{l}, \bm{x}_{l-1}, \bm{o}^s_{T}).\!\!
\end{aligned}
\vspace{-2pt}
\end{equation}
Finally, we have $P(X|A; E)\!=\!\prod_lp_l(x_l)$, where $p_l(x_l)\!=\!P(x_l|x_{1:l-1}, a_{1:T},_{\!} V_{1:T\!})$$_{\!}$ for$_{\!}$ brevity.$_{\!}$ For$_{\!}$ fair$_{\!}$ comparison,$_{\!}$ our$_{\!}$ speaker is the same as, but not specific to, the one in~\cite{fried2018speaker}.

\noindent\textbf{Cycle-Consistent Training.} Given aligned $(E, X)\!\in\!\mathcal{E}\!\times\!\mathcal{X}$, we can first obtain a navigation path $\hat{A}$ through the follower $f(E, X)$ (shortly $f_E(X)$). We then use the speaker to translate $\hat{A}$ to a visual-grounded instruction $s(E, \hat{A})$ (shortly $s_E(\hat{A})$), which is expected to be semantically similar to $X$, \ie, gaining a small  cycle-consistent error $\triangle_E(X, s_E(\hat{A}))$ (shortly $\triangle^X_E$). Similarly, for paired $(E, A)\!\in\!\mathcal{E}\!\times\!\mathcal{A}$, we have $\triangle^A_E$. Then the $\triangle$ error can be specified as the negative log-likelihood, which is minimized for regularizing training:
\vspace{-2pt}
\begin{equation}\small\label{eq:6}
\begin{aligned}
\!\!\!\!\!\!\triangle^A_E\!=\!&-\!\log\!\!\sum_{\hat{X}\in\mathcal{X}} \!\!P(s_E(A)\!=_{\!}\!\hat{X}|A)P(f_E(\hat{X})\!=_{\!}\!A|s_E(A)\!=_{\!}\!\hat{X}),\!\!\!\\
\!\!\!\!\!\!\triangle^X_E\!=\!&-\!\log\!\!\sum_{\hat{A}\in\mathcal{A}} \!\!P(f_E(X)\!=_{\!}\!\hat{A}|X)P(s_E(\hat{A})\!=_{\!}\!X|f_E(X)\!=_{\!}\!\hat{A}).\!\!\!
\end{aligned}
\vspace{-3pt}
\end{equation}
For ease of reference, we briefly denote Eq.~\ref{eq:6}  as:
\vspace{-3pt}
\begin{equation}\small\label{eq:7}
\begin{aligned}
\!\!\!\triangle^A_E\!=\!&-\!\log\sum\nolimits_{\hat{X}\in\mathcal{X}} P(\hat{X}|A; E)P(A|\hat{X}; E),\!\!\!\\
\!\!\!\triangle^X_E\!=\!&-\!\log\sum\nolimits_{\hat{A}\in\mathcal{A}} P(\hat{A}|X; E)P(X|\hat{A}; E).
\end{aligned}
\vspace{-3pt}
\end{equation}
Directly calculating the gradients for $\triangle^A_E$ is intractable due to the huge space of $\mathcal{X}$. A similar issue also holds for  $\triangle^X_E$. Inspired by~\cite{he2016dual,wang2019multi}, the gradients for $\triangle^A_E$ \wrt the parameters of the follower and speaker, \ie, $\theta^f$ and $\theta^s$, can be computed as (and similarly for $\triangle^X_E$):
\vspace{-3pt}
\begin{equation}\small\label{eq:8}
\begin{aligned}
\!\!\!\frac{\partial\triangle^A_E}{\partial{\theta^f}}&\!\approx\!-\mathbb{E}_{\hat{X}\sim P(\cdot|A; E)}[\frac{\partial\log P(A|\hat{X};E)}{\partial{\theta^f}}]\\&\!\approx\!-\frac{\partial\log P(A|\hat{X};E)}{\partial{\theta^f}},\\
\!\!\!\frac{\partial\triangle^A_E}{\partial{\theta^s}}&\!\approx\!-\mathbb{E}_{\hat{X}\sim P(\cdot|A; E)}[\log P(A|\hat{X};E)\frac{\partial\log P(\hat{X}|A; E)}{\partial{\theta^s}}]\\
&\!\approx\!-(\log P(A|\hat{X};E)-b^f)\frac{\partial\log P(\hat{X}|A;E)}{\partial{\theta^s}},
\end{aligned}
\vspace{-3pt}
\end{equation}
where $b^f$ is a baseline to reduce the training variance and estimated by the mean of previous $\log P(A|\hat{X}; E)$. Please see our supplementary for more details.

Hence, as the estimation for the $\triangle$ error does not require any aligned instruction-trajectory pairs $(X, A)$, Eq.~\ref{eq:6} can be naturally applied over both labeled and unlabeled paths for training. Let us denote $\mathcal{D}\!=\!\{(E_n, X_n, A_n)\}_{n=1}^N$ as a labeled data collection which consists of $N$ aligned environment-instruction-path$_{\!}$ tuples. As in conventions~\cite{fried2018speaker,tan2019learning,ke2019tactical}, we can also build an unlabeled data collection, $\mathcal{U}\!=$ $\!\{(E_m, {A}'_m)\}_{m=1}^M$ through sampling some paths ${A}'_m$ from existing environments $E_m$ (but without instruction annotations). Then our cycle-consistent learning loss is defined as:
\vspace{-7pt}
\begin{equation}\small\label{eq:9}
\begin{aligned}
\mathcal{L}_{cycle}\!=\!\frac{1}{N}\!\sum_{(E,X,A)~\!\in~\!\mathcal{D}}\!\!(\triangle^A_E\!+\!\triangle^X_E)+\frac{1}{M}\!\sum_{(E, {A}')~\!\in~\!\mathcal{U}}\!\!\triangle^{A'}_E.
\end{aligned}
\vspace{0pt}
\end{equation}
Other learning objectives for instruction following and generation, defined over the labeled triplets $(E, X, A)$, are also compatible and used in our training stage (\textit{cf}.~\S\ref{sec:DNA}).

\noindent\textbf{Remark.}  Our cycle-consistency design is driven by two
beliefs. First, a desired VLN agent should be able to ground both navigation actions and linguistic cues together on the visual environments. Thus it is necessary to explore both navigation planning and instruction creation in a unified learning scheme, enabling the agent to better capture cross-modal and cross-task connections. Second, assuming the generated instruction $s_E(A)$ is a valid rephrasing of the original
$X$, a robust follower should execute this rephrasing $s_E(A)$ with the same navigation plan as the original $X$. A similar conclusion also holds for the speaker.
\vspace{-2pt}
\subsection{Counterfactual Environment Creation}\label{sec:CEC}
\vspace{-2pt}
Back-translation based data augmentation~\cite{fried2018speaker} has become a common practice in VLN. The central idea is to translate sampled paths into artificial instructions and use these synthesized environment-instruction-path tuples to augment the labeled data $\mathcal{D}$. Aside from learning the speaker and follower in isolation, it involves \textit{multiple} training stages (\textit{cf}.~\S\ref{sec:1}). Contrarily, our cycle-consistent learning, conducted on both labeled $\mathcal{D}$ and unlabeled  data $\mathcal{U}$, has a unified training objective (\textit{cf.}~Eq.~\ref{eq:9}). Interestingly, recent studies~\cite{fu2019counterfactual} suggested the advantage of such path sampling strategy~\cite{fried2018speaker}  in counterfactual thinking: a path $A'$, sampled in $E$, and its artificial instruction $\hat{X}'$ compose a counterfactual. Although our cycle-consistent learning scheme has naturally involved such kind of counterfactuals during the computation of $\triangle^{A'}_E$ in Eq.~\ref{eq:9}, to fully explore the potential of counterfactual thinking, we further propose an environment creator $c$ that generates counterfactual observations by greatly changing house layouts but without interfering with the execution of original instructions.

\noindent\textbf{Creator.} Our goal is to learn a creator $c\!:\! (E, X, A)\!\mapsto\!\bar{E}$ that observes the environment $E$, instruction $X$, navigation path $A$ and generates a counterfactual environment $\bar{E}$ such that i) the differences between $E$ and $\bar{E}$ are large; and ii) $\bar{E}$ and $(X, A)$ present high compatibility. Such design is based upon previous research that has proved that,  i) human prefer large modifications (even introducing elements not present in the real experience), during counterfactual thinking~\cite{girotto2007postdecisional}; and, ii) the imagined world should not be completely divorced form the reality (with proper changes)~\cite{woodward2011psychological}.

For ease of optimization, instead of only considering the aligned triplet $(E, X, A)\!\in\!\mathcal{D}$, the creator $c$ additionally uses other real scene $E^r$ (sampled from $\mathcal{D}$) as the reference. By mixing $E$ with $E^r$, it creates a counterfactual environment $\bar{E}$,  under the above-mentioned constraints: i) \textbf{diversity}: replacing elements in $E$, as many as possible, with the ones in $E^r$; ii) \textbf{realism}: maintaining $(X, A)$ still feasible in the changed environment $\bar{E}$. To do so, a compact descriptor $\bm{u}$ is first generated for $(E, X, A)$:\!\!
\vspace{-3pt}
\begin{equation}\small\label{eq:10}
\begin{aligned}
\bm{u}\!=\!\bm{h}^c_T, ~~~~\bm{h}^c_t = \text{LSTM}^c([\bm{V}_{t}, {\bm{a}}_t, \bm{X}], \bm{h}^c_{t-1}).
\end{aligned}
\vspace{-3pt}
\end{equation}
Here $\bm{u}$ is desired to encode all the necessary information that make  $(E, X, A)$ valid. As shown in Fig.~\ref{fig:method}(b), given an observed scene $V\!\in\!E$ and its reference $V^r\!\in\!E^r$, the creator $c$ fuses them together, in the feature rather than pixel space:\!\!
\vspace{-3pt}
\begin{equation}\small\label{eq:11}
\begin{aligned}
\bm{q}_k &= \text{softmax}([\bm{v}^r_1, \bm{v}^r_2, \cdots, \bm{v}^r_K]^{\top\!}\cdot\bm{v}_k),\\
\bm{g}_k &= [\bm{v}^r_1, \bm{v}^r_2, \cdots, \bm{v}^r_K]\cdot\bm{q}_k,\\
\lambda_k &= \text{sigmoid}(\bm{u}^{\!\top\!}\bm{W}_3\bm{v}_k),\\
\bar{\bm{v}}_k &= \lambda_k\bm{v}_k+(1-\lambda_k)\bm{g}_k,
\end{aligned}
\vspace{-3pt}
\end{equation}
where $\bm{v}_k$ is the embedding of a visual element~${v}_k$~in$_{\!}$ scene$_{\!}$ $V$ (\ie, $\bm{V}_{\!}\!=\!\text{max-pool}([\bm{v}_1, \bm{v}_2,_{\!} \cdots_{\!}, \bm{v}_K]$)), $\bm{q}_k$ refers to normalized correlation score vector between ${v}_k$ and ${V}^{r\!}\!=\!\{v^r_k\}_{k}$, $\bm{g}_k$ indicates the attention summary, and $\lambda_k$ gives the importance of $\bm{v}_k$ for the successful execution of $(E, X, A)$~and decides whether $\bm{v}_k$ needs to be replaced. Eventually, we have $\bar{\bm{v}}_k$, \ie, the embedding of a visual region~$\bar{v}_k$~in$_{\!}$ the created counterfactual scene$_{\!}$ $\bar{V}\in\bar{E}$.

\noindent\textbf{Training$_{\!}$ Objective.}$_{\!}$ With$_{\!}$ the$_{\!}$ purposes$_{\!}$ of$_{\!}$ i)$_{\!}$ modifying$_{\!}$ the original scene$_{\!}$ $V$ as$_{\!}$ much$_{\!}$ as$_{\!}$ possible and$_{\!}$ ii)$_{\!}$ keeping$_{\!}$ crucial information/landmarks aligned$_{\!}$ with$_{\!}$ the$_{\!}$ instruction-path$_{\!}$ pair $(X, E)$, the training loss of the creator$_{\!}$ $c$ is$_{\!}$ designed$_{\!}$ as:
\vspace{-3pt}
\begin{equation}\small\label{eq:12}
\begin{aligned}
\mathcal{L}^c&=\mathcal{L}_{\ell_2}+\mathcal{L}_{adv},\\
             &=||\bm{\lambda}||_2+\log(1-d(\bar{V}, X, A)).
\end{aligned}
\vspace{-3pt}
\end{equation}
The$_{\!}$ $\ell_2$-norm$_{\!}$ loss$_{\!}$ $\mathcal{L}_{\ell_2\!}$ inspires$_{\!}$ the$_{\!}$ sparsity$_{\!}$ of$_{\!}$ $\bm{\lambda}\!=\![\lambda_k]_k$~and hence$_{\!}$ addresses$_{\!}$ i). The$_{\!}$ adversarial$_{\!}$ loss$_{\!}$ $\mathcal{L}_{adv}$$_{\!}$ tries to ``fool''$_{\!}$ a$_{\!}$ discriminator$_{\!}$ $d\!:\! \mathcal{E}_{\!}\times_{\!}\mathcal{X}_{\!}\times_{\!}\mathcal{A}\!\mapsto\![0,1]$.$_{\!}$ The$_{\!}$ discriminator $d$ is learnt to estimate the alignment between environments and instruction-action pairs by minimizing $d(\bar{V}, X, A)_{\!}-_{\!}d(V_{\!}, X, A)$. Thus $\mathcal{L}_{adv}$ addresses ii). Note that $d$ only uses geometric action embedding. Optimizing above objectives makes $(\bar{E}, X, A)$ a valid training example. Therefore, in our counterfactual environment $\bar{E}$, we can still train the speaker to translate the navigation path $E$ into $X$, and train the follower to execute the instruction $X$ as $A$.

\begin{figure*}
\begin{minipage}{\textwidth}
\begin{minipage}[t]{0.295\textwidth}
   	\vspace*{-2pt}
\subsection{Implementation Details}\label{sec:DNA}
	\vspace*{-2pt}
\noindent\textbf{Network Architecture.} We implement our \textbf{follower} $f$ with different architectures\!~\cite{anderson2018vision,fried2018speaker,wang2019reinforced}. Instruction embedding $\bm{{X}}_{\!}$ is from an LSTM based linguistic encoder as normally. With~\cite{anderson2018vision}, only front view is used and corresponding embedding $\bm{V}$ is obtained from a pretrained ResNet-152~\cite{he2016deep}. Action embedding $\bm{a}$ is also from the linguistic LSTM. With \cite{fried2018speaker,wang2019reinforced}, the panoramic view is perceived and divided into 36 sub-views
   \end{minipage}
   \begin{minipage}[t]{0.005\textwidth}
   ~~~~~~
   \end{minipage}
 \begin{minipage}[t]{0.7\textwidth}
    \begin{threeparttable}
        \resizebox{0.98\textwidth}{!}{
		\setlength\tabcolsep{4pt}
		\renewcommand\arraystretch{1.0}
\begin{tabular}{c||cccc|cccc|cccc}
\hline 
~ &  \multicolumn{4}{c|}{\texttt{val} \texttt{seen}} & \multicolumn{4}{c|}{\texttt{val} \texttt{unseen}} & \multicolumn{4}{c}{\texttt{test} \texttt{unseen}} \\
\cline{2-13}
\multirow{-2}{*}{Model} &\textbf{SR}$\uparrow$ &NE$\downarrow$ &OR$\uparrow$  &SPL$\uparrow$ &\textbf{SR}$\uparrow$ &NE$\downarrow$  &OR$\uparrow$  &SPL$\uparrow$ &\textbf{SR}$\uparrow$ &NE$\downarrow$ &OR$\uparrow$ &SPL$\uparrow$\\
\hline
\hline
\textit{Seq2Seq}~\cite{anderson2018vision} & 39.4 & 6.0  & 51.7  & 33.8   &  22.1 & 7.8  &27.7  &19.1 &  20.4 & 7.9  & 26.6   &18.0\\
~~~~~+ {BT}~\cite{fried2018speaker}  &43.7 &5.3 & 58.1&37.2   &22.6 & 7.7  &28.9 &19.9 &21.0 &7.8&26.2&18.8\\
~~~~~~~+ {APS}~\cite{fu2019counterfactual} &48.2 & \textbf{5.0} & 60.8&40.1   &24.2 & 7.1& 32.7 &20.4&22.5&\textbf{7.5}&30.1&19.3\\
+ {CCC} & \textbf{50.1} & \textbf{5.0}  & \textbf{61.1} & \textbf{42.6} & \textbf{28.4} & \textbf{6.8}  & \textbf{35.3} & \textbf{22.1} & \textbf{25.5} & 7.8 & \textbf{35.9} & \textbf{20.6}\\
\hline
\textit{Speaker-Follower}~\cite{fried2018speaker}& 51.7 & 5.0 &61.6 &44.4  &29.9 &6.9  & 40.7 & 21.0 &30.9 &  7.0 & 41.2&24.0\\
~~~~~+ {BT}~\cite{fried2018speaker}  & 66.4 & 3.7  & 74.2 & 59.8                & 36.1 &6.6  &46.6  & 28.8 & 34.8 & 6.6  & 43.4 & 29.2\\
~~~~~~~+ {APS}~\cite{fu2019counterfactual} & 68.2 & \textbf{3.3}  & \textbf{74.9} &\textbf{62.5}               &38.8 &6.1  &46.7 & 32.1 &36.1 &6.5 &44.2 &28.8\\
+ {CCC} & \textbf{68.4} & \textbf{3.3} & {74.5} & {61.4} & \textbf{43.5} & \textbf{5.8} & \textbf{52.0} &\textbf{38.1} & \textbf{41.4} & \textbf{5.9} & \textbf{51.0} & \textbf{36.6} \\
\hline
\textit{RCM}~\cite{wang2019reinforced} &47.0 &5.7  &53.8 &44.3     &35.0 & 6.8  &43.0 &31.4                     &35.9 &6.7 &43.5& 33.1\\
~~~~~+ {BT}~\cite{fried2018speaker} & 61.9 &4.1  & 66.9 &58.6   &45.6 &5.7 &  52.4 &41.8   &44.5 &5.9 & 52.4 &40.8\\
~~~~~~~+ {APS}~\cite{fu2019counterfactual} &63.2 &3.9 & 69.3 &59.5   & 47.7 & 5.4 &56.6 & 42.8                    &45.1 &5.8 & 53.9&40.9\\
+ {CCC} & \textbf{68.0} & \textbf{3.4}  & \textbf{77.5} & \textbf{62.1} & \textbf{50.4} & \textbf{5.2}  & \textbf{57.8} & \textbf{46.4} & \textbf{51.0} & \textbf{5.3} & \textbf{57.2} & \textbf{48.2}\\\hline
\end{tabular}
}
\end{threeparttable}
    \vspace*{-8pt}
    \makeatletter\def\@captype{table}\makeatother\captionsetup{font=small}\caption{\small{Quantitative comparison results (\S\ref{sec:ex_if}) for instruction following on R2R dataset~\cite{anderson2018vision}.  }
    \label{table:R2Rif}}
  \end{minipage}
  \end{minipage}
  \vspace*{-20pt}
\end{figure*}

\noindent\textbf{Remark.} The creator is fully differentiable and trained with the speaker and follower together, leading to a triple-agent learning system. During training, the cycle-consistent errors, \ie, $\triangle^A_{\bar{E}}$ and $\triangle^X_{\bar{E}}$, can also be estimated and minimized over the counterfactual samples, \ie, $(\bar{E}, X, A)$, generated by the creator. Moreover, the creator can also access the supervision signals of cycle-consistent learning (\textit{cf.}~Eq.~\ref{eq:9}) and training objectives for instruction following and generation learning (detailed in \S\ref{sec:DNA}). Thus the creator can
progressively create more informative counterfactuals on-the-fly, in turn boosting the training of the speaker and follower.

\noindent(12\!~headings\!~$\times$\!~3\!~elevations with $30^\circ$ intervals). Each sub-view is associated with a geometric feature, \ie, $(\cos\phi_h,$ $ \sin\phi_h,$ $\cos\phi_e,$ $\sin\phi_e)$, where $\phi_h$ and $\phi_e$ are the angles of heading and elevation, respectively. Visual and geometric features are concatenated as the embedding of the sub-view and corresponding action. See~\cite{anderson2018vision,fried2018speaker,wang2019reinforced} for more network details. The model design of our \textbf{speaker} $s$ follows the one in~\cite{fried2018speaker}, which is built upon the panoramic view system. For the \textbf{creator} $c$, the discriminator $d$ only uses geometric information based action representation (to filter out the visual cues in trajectories). Both $c$ and  $d$ adopt cross-modal co-attention based network architectures, like~\cite{tan2019learning}.

\noindent\textbf{Training.} Besides minimizing the cycle-consistent loss $\mathcal{L}_{cycle\!}$ (\textit{cf.}~Eq.~\ref{eq:9}), our CCC framework is also learnt with the training objectives for instruction generation and following, over the labeled data  $\mathcal{D}$ and counterfactual samples. For instruction following, IL~\cite{anderson2018vision} is adopted for off-policy learning, where the loss is defined over the groundtruth navigation action sequence $A$: $\mathcal{L}_{IL\!}^{f}\!=\!-\log P(A|X)$.  RL is also applied for on-policy learning \cite{wang2019reinforced,tan2019learning}, \ie, optimizing $\mathcal{L}_{RL\!}^{f}\!=\!-\sum_t\log p_t(a_t)\Lambda_t$, where $a_t\!\sim\!p_t(a_t)$ and $\Lambda_t$ indicates the advantage in A2C~\cite{mnih2016asynchronous}. For instruction generation, the speaker is trained with $\mathcal{L}^{s\!}\!=\!-\log P(X|A)$, where $X$ refers to the groundtruth navigation instruction. To stabilize training, we apply an annealing strategy~\cite{tan2019learning} to the IL signal which makes the agents learn a good initial policy.

\noindent\textbf{Inference.} Once trained, the speaker and follower can perform their specific tasks independently. As in conventions, we apply greedy prediction, \ie, $x^*_l\!=\!\arg\max(p_l(x_l))$ and $a^*_t\!=\!\arg\max(p_t(a_t))$, for instruction creation and following, as approximations of $X^*\!=\!\arg\max P(X|A; E)$ and $A^*\!=\!\arg\max P(A|X; E)$, respectively.

	\vspace*{-4pt}
\section{Experiment}\label{sec:ex}

	\vspace*{-2pt}
\subsection{Performance on Instruction Following}\label{sec:ex_if}
	\vspace*{-2pt}
\noindent\textbf{Dataset.}~We conduct experiments on R2R~\cite{anderson2018vision}, originally developed for the instruction following task. R2R has four sets: \texttt{train} ($61$ environments, $14,039$ instructions), \texttt{val} \texttt{seen} ($61$ environments, $1,021$
instructions), \texttt{val} \texttt{unseen} ($11$ environments, $2,349$ instructions), and \texttt{test} \texttt{unseen} ($18$ environments, $4,173$ instructions). There are no overlapping environments between the unseen and training sets.

\noindent\textbf{Evaluation Metric.} Following~\cite{anderson2018vision,fried2018speaker}, four standard metrics for instruction following are used:
1) \textit{Success rate} (SR) computes the percentage of final positions less than 3 m away$_{\!}$ from$_{\!}$ the$_{\!}$ goal$_{\!}$ location.$_{\!}$
2)$_{\!}$ \textit{Navigation$_{\!}$ error}$_{\!}$ (NE) refers to the shortest distance between agent's final position and the goal location.
3) \textit{Oracle success rate} (OR) is the success rate if the agent can stop at the closest point to the goal along its trajectory.
4) \textit{Success rate weighted by path length} (SPL)~\cite{anderson2018evaluation}  is a trade-off between  SR and navigation length.

\begin{table*}[!b]
	\centering
	\vspace*{-6pt}
	  \renewcommand\thetable{3}
			\resizebox{1\textwidth}{!}{
			\setlength\tabcolsep{4pt}
			\renewcommand\arraystretch{1.0}
	\begin{tabular}{c||cccccc|cccccc}
	\hline 
	~ &  \multicolumn{6}{c|}{\texttt{val} \texttt{seen}} & \multicolumn{6}{c}{\texttt{val} \texttt{unseen}}\\
	\cline{2-13}
	\multirow{-2}{*}{Model}
	&Bleu-1~$\uparrow$ &Bleu-4~$\uparrow$ &CIDEr~$\uparrow$ &Meteor~$\uparrow$ &Rouge~$\uparrow$ &\textbf{SPICE}~$\uparrow$ &Bleu-1~$\uparrow$ &Bleu-4~$\uparrow$ &CIDEr~$\uparrow$ &Meteor~$\uparrow$ &Rouge~$\uparrow$ &\textbf{SPICE}~$\uparrow$ \\
	\hline
	\hline
	BT Speaker~\cite{fried2018speaker} &0.537 &0.155 &0.121 &0.233 &0.350 &0.203  &0.522 &0.142 &0.114 &0.228 &0.346 &0.188 \\
	VLS~\cite{agarwal2019visual} &0.549 &0.157 &0.137 &0.228 &0.352 &0.214  &0.548 &0.159 &0.132 &0.231 &0.357 &0.197 \\
	Ours-\textit{Seq2Seq} & 0.720 & 0.296 & 0.529 & 0.233 & 0.487 & 0.216 & 0.704 & 0.273 & 0.475 & 0.229 & 0.473 & 0.202\\
	Ours-\textit{Speaker-Follower} & 0.723 & \textbf{0.299} & \textbf{0.566} & 0.235 & 0.490 & 0.229 & 0.706 & \textbf{0.275} & \textbf{0.477} & 0.229 & 0.474 & 0.207 \\
	Ours-\textit{RCM} &\textbf{0.728} & 0.287 & 0.543 & \textbf{0.236} & \textbf{0.493} & \textbf{0.231} & \textbf{0.708} & 0.272 & 0.461 & \textbf{0.231} &\textbf{ 0.477} & \textbf{0.214}\\
	\hline
	\end{tabular}
	}
		\vspace*{-2pt}
	\captionsetup{font=small}
		\caption{\small{Quantitative comparison results (\S\ref{sec:ex_ig}) for instruction generation on R2R dataset~\cite{anderson2018vision}.  }}
		\label{table:R2Rig}
	\vspace*{-2pt}
\end{table*}

\noindent\textbf{Evaluation Protocol.} As in~\cite{fried2018speaker,fu2019counterfactual}, we test our model with several representative baselines~\cite{anderson2018vision,fried2018speaker,wang2019reinforced} using different architectures, action spaces, and learning paradigms.
 \begin{itemize}[leftmargin=*]
	\setlength{\itemsep}{0pt}
	\setlength{\parsep}{1pt}
	\setlength{\parskip}{1pt}
	\setlength{\leftmargin}{-10pt}
	\vspace{-6pt}
	\item \textit{Seq2Seq}~\cite{anderson2018vision}: an attention-based {\texttt{Seq2Seq}} model that is trained with IL under the visuomotor action space.
	\item \textit{Speaker-Follower}~\cite{fried2018speaker}: a compositional model that is trained with IL under the panoramic action space.
	\item \textit{RCM}\!~\cite{wang2019reinforced}:$_{\!}$ an$_{\!}$ improved$_{\!}$ multi-modal$_{\!}$ model$_{\!}$ that$_{\!}$ is$_{\!}$ trained$_{\!}$ using$_{\!}$ both$_{\!}$ IL and RL under the panoramic action space.
	\vspace{-5pt}
\end{itemize}
The baselines are trained with labeled instruction-path pairs$_{\!}$ in$_{\!}$ R2R$_{\!}$ \texttt{train}$_{\!}$ set.$_{\!}$ For$_{\!}$ each$_{\!}$ baseline,$_{\!}$ we further report the performance with our CCC and other two speaker-based data augmentation techniques, \ie, back-translation (BT)~\cite{fried2018speaker} and adversarial path sampling (APS) \cite{fu2019counterfactual}:
\begin{itemize}[leftmargin=*]
	\setlength{\itemsep}{0pt}
	\setlength{\parsep}{1pt}
	\setlength{\parskip}{1pt}
	\setlength{\leftmargin}{-10pt}
	\vspace{-5pt}
	\item {CCC}: our$_{\!}$ speaker$_{\!}$ is$_{\!}$ jointly$_{\!}$ learnt$_{\!}$ with$_{\!}$ the$_{\!}$ follower$_{\!}$ in$_{\!}$ real$_{\!}$ and counterfactual environments and translates randomly sampled paths into instructions as extra training data.
	\item {BT}~\cite{fried2018speaker}: the speaker is trained in isolation with the follower in real environments only and translates randomly sampled paths into instructions as extra training data.
	\item {APS}~\cite{fu2019counterfactual}: the speaker is trained in isolation with the follower in real environments only but translates adversarially selected paths into instructions as$_{\!}$ extra training data.$_{\!}$
	\vspace{-15pt}
\end{itemize}

\begin{table}[t]
	\centering
	  \renewcommand\thetable{2}
			\resizebox{0.49\textwidth}{!}{
			\setlength\tabcolsep{8pt}
			\renewcommand\arraystretch{1.0}
\hspace{-1.0em}
	\begin{tabular}{c||cccc}
	\hline 
	~ & \multicolumn{4}{c}{\texttt{test} \texttt{unseen}} \\
	\cline{2-5}
	\multirow{-2}{*}{Models} &\textbf{SR}$\uparrow$ &NE$\downarrow$ &OR$\uparrow$ &SPL$\uparrow$\\
	\hline
	\hline
	Self-Monitoring~\cite{ma2019self}& 43.0 & 6.0 & 55.0 & 32.0 \\
	Regretful~\cite{ma2019regretful} & 48.0 & 5.7  & 56.0  & 40.0 \\
	OAAM~\cite{qi2020object}&53.0  &-   &61.0  & 50.0 \\
	Tactical Rewind~\cite{ke2019tactical}&54.0 & 5.1  & 64.0 & 41.0 \\
	AuxRN~\cite{zhu2019vision}& 55.0 &5.2  &62.0 & \textbf{51.0} \\
	\hline
	E-Dropout~\cite{tan2019learning} & 48.0 & 5.6  & 58.0  & 44.0 \\
	E-Dropout~\cite{tan2019learning} + CCC & 52.2 & 5.1  & 59.8  & 46.9 \\
	Active Perception~\cite{wang2020active} &55.7 & 4.8 & \textbf{73.1} & 37.1\\
	Active Perception~\cite{wang2020active} + CCC & 60.6 & 4.3 & 71.4 & 41.3\\
	{SSM}~\cite{wang2021structured} & 57.3 & 4.7  & 68.2 & 44.1 \\
	{SSM}~\cite{wang2021structured} + CCC & \textbf{62.2} & \textbf{4.3} & 72.3 & 49.2 \\
	\hline
	\end{tabular}
	}
		\vspace*{-2pt}
	\captionsetup{font=small}
    \captionsetup{width=.99\linewidth}
		\caption{\small{Benchmarking results (\S\ref{sec:ex_if}) for instruction following on R2R dataset~\cite{anderson2018vision}. }}
		\label{table:R2R}
	\vspace*{-10pt}
\end{table}

\begin{table*}[t]
		\centering
				\resizebox{0.97\textwidth}{!}{
				\setlength\tabcolsep{5pt}
				\renewcommand\arraystretch{1.05}
		\begin{tabular}{c|c||cccc|cccccc}
		\hline 
		~ & & \multicolumn{4}{c|}{Instruction Following} & \multicolumn{6}{c}{Instruction Generation}\\
		\cline{3-12}
		\multirow{-2}{*}{Model} &\multirow{-2}{*}{Component}
		&\textbf{SR}$\uparrow$ &NE$\downarrow$ &OR$\uparrow$  &SPL$\uparrow$ &Bleu-1~$\uparrow$ &Bleu-4~$\uparrow$ &CIDEr~$\uparrow$ &Meteor~$\uparrow$ &Rouge~$\uparrow$ &\textbf{SPICE}~$\uparrow$ \\
		\hline
		\hline
		Baseline~\cite{fried2018speaker} &- &29.9 &6.9  & 40.7 & 21.0 &0.522 &0.142 &0.114 &0.228 &0.346 &0.188 \\
		\hline
		\multirow{4}{*}{\tabincell{c}{Cycle-\\Consistency}}
		&$\triangle^A_E$& 33.2 & 6.7 & 44.7 & 23.7 & 0.694 & 0.269 & 0.446 & 0.228 & 0.471 & 0.192\\
		&$\triangle^X_E$& 28.6 & 6.9  & 40.1 & 20.5 & 0.699 & 0.271 & 0.456 & 0.228 & 0.472 & 0.195\\
		&$\triangle^A_E\!+\!\triangle^X_E$& 35.1 & 6.7 & 44.5 & 25.4 & 0.702 & 0.272 & 0.462 & 0.229 & 0.472 & 0.196\\
		&$\triangle^A_E\!+\!\triangle^X_E\!+\!\triangle^{A'}_E$ & 37.7 & 6.6 & 45.9 & 26.7 & 0.703 & 0.272 & 0.467 & 0.229 & 0.471 & 0.198\\
		\hline
		\multirow{2}{*}{\tabincell{c}{Counterfactual\\Environment}} & \textit{w/o} reference environment $E^r$ &29.9 &6.9  & 40.7 & 21.0 &0.522 &0.142 &0.114 &0.228 &0.346 &0.188\\
		& \textit{w} reference environment $E^r$ & 41.7 & 5.9 & 50.7 & 35.6 & 0.701 & 0.271 & 0.459 & 0.229 & 0.472 & 0.200\\
		\hline
		Full Model &$\triangle^A_E\!+\!\triangle^X_E\!+\!\triangle^{A'}_E\!+\!\triangle^A_{\bar{E}}\!+\!\triangle^X_{\bar{E}}$ & 43.5 & 5.8 & 52.0 &38.1 & 0.706 & 0.275 & 0.477 & 0.229 & 0.474 & 0.207
		\\
		\hline
		\end{tabular}
		}
			\vspace*{0pt}
		\captionsetup{font=small}
			\caption{\small{Ablation study on \texttt{val} \texttt{unseen} of R2R dataset~\cite{anderson2018vision}. See \S\ref{sec:abl} for details.}}
			\label{table:abs}
		\vspace*{-8pt}
\end{table*}

\noindent\textbf{Quantitative Result.} The comparison results on instruction following are summarized in Table~\ref{table:R2Rif}. We find CCC outperforms other learning paradigms across diverse dataset splits and metrics. For the three baseline followers~\cite{anderson2018vision,fried2018speaker,wang2019reinforced}, CCC gains remarkable SR improvements (\ie, 0.2-4.8, 2.7-4.7, and 3.0-6.1) compared with the second best on \texttt{val}~\texttt{seen}, \texttt{val}~\texttt{unseen} and \texttt{test} \texttt{unseen} respectively. This validates the efficacy of CCC across different follower architectures. Further, the performance improvements on \texttt{unseen} sets are relatively more significant, showing that CCC strengthens$_{\!}$ models'$_{\!}$ generalization ability.

\noindent\textbf{Performance Benchmarking.}  For comprehensive evaluation, we conduct performance benchmarking by applying our CCC technique to~\cite{tan2019learning,wang2021structured,wang2020active}, which are current top-leading  instruction followers with public implementations:
\begin{itemize}[leftmargin=*]
	\setlength{\itemsep}{0pt}
	\setlength{\parsep}{1pt}
	\setlength{\parskip}{1pt}
	\setlength{\leftmargin}{-10pt}
	\vspace{-5pt}
	\item \textit{E-Dropout}~\cite{tan2019learning}: a multi-modal model that uses `environmental dropout' method to mimic unseen environments.
	\item {\textit{Active Perception}~\cite{wang2020active}: a robust model that  is able to actively explore surroundings for more intelligent planning.}
    \item \textit{SSM}~\cite{wang2021structured}: a graph model which is equipped with a map building module for global decision making.
	\vspace{-5pt}
\end{itemize}

As shown in Table~\ref{table:R2R}, our CCC greatly boosts the performance of current three top-performing instruction followers \cite{tan2019learning,wang2021structured,wang2020active}, across all the metrics. For example, SR and SPL of E-Dropout~\cite{tan2019learning} are improved by 4.2 and 2.9, respectively. For Active Perception~\cite{wang2020active}, it obtains significant performance gains, \eg, 4.9 SR and 4.2 SPL. Based on SSM~\cite{wang2021structured}, our CCC further improves the SR to 62.2.

\subsection{Performance on Instruction Generation}\label{sec:ex_ig}
\vspace{-2pt}
\noindent\textbf{Dataset.}~As R2R~\cite{anderson2018vision} \texttt{test} \texttt{unseen} set is preserved for benchmarking instruction following methods, we report the performance of instruction generation on \texttt{val} \texttt{seen} and \texttt{val} \texttt{unseen} sets. Note that, in R2R, each path is associated with three ground-truth navigation instructions.

\begin{figure*}[!tbh]
  \centering
      \includegraphics[width=0.99\linewidth]{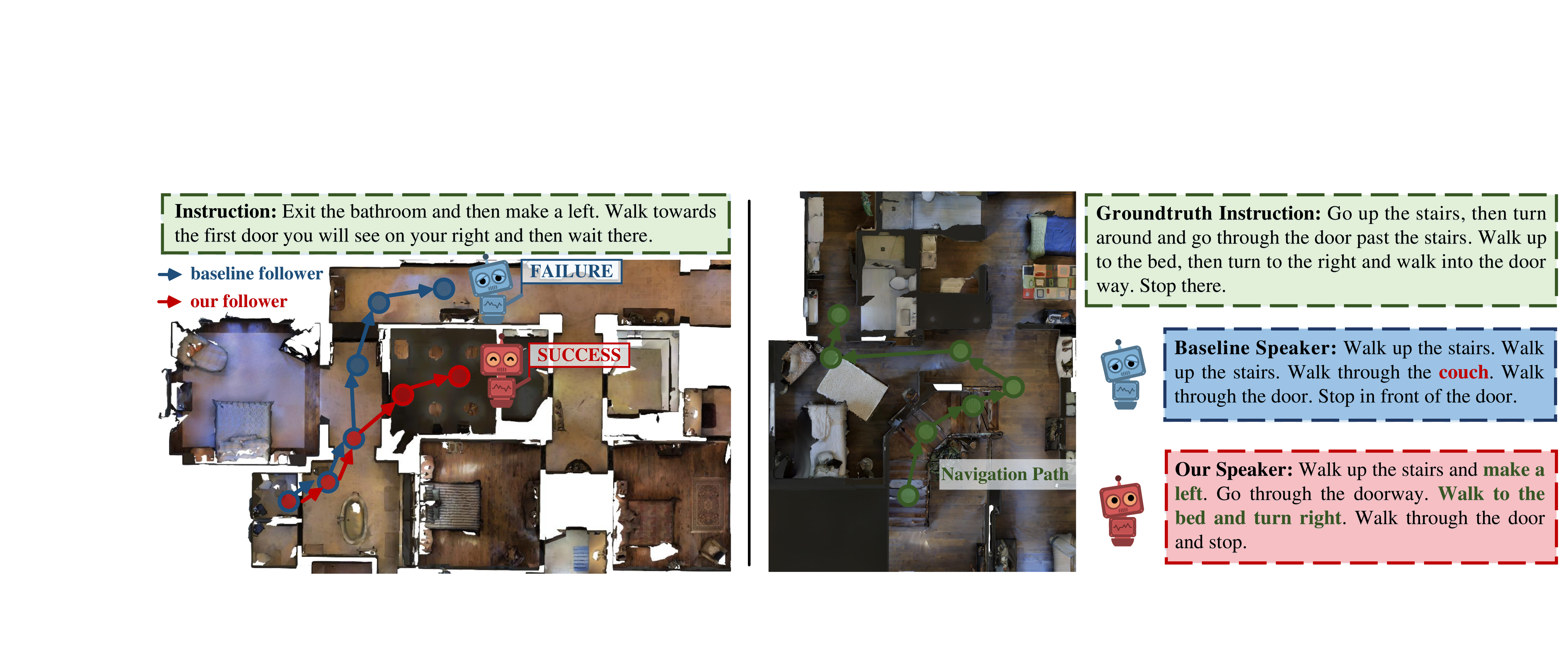}
\vspace{-5pt}
\captionsetup{font=small}
	\caption{\small{Visual comparison results on instruction following (left) and instruction  generation (right).  See \S\ref{sec:abl} for details.
}
}
\label{fig:vr}
\vspace{-13pt}
\end{figure*}

\noindent\textbf{Evaluation Metric.} Five standard textual evaluation metrics are considered here~\cite{agarwal2019visual}: 1) BLEU~\cite{papineni2002bleu} refers to the geometric mean of $n$-gram precision scores computed over reference and candidate descriptions. 2) CIDEr~\cite{vedantam2015cider} first represents each
sentence with a set of 1-4 grams and calculates the co-occurrences of $n$-grams in the reference sentences and candidate sentence as the score. 3) METEOR~\cite{banerjee2005meteor} is defined as the
harmonic mean of precision and recall of unigram matches between sentences.  4) ROUGE~\cite{lin2004rouge} is computed by comparing overlapping $n$-grams, word sequences and word pairs. 5) SPICE~\cite{anderson2016spice} is based on the agreement of the scenegraph~\cite{schuster2015generating} tuples  of the candidate sentence and all reference sentences. These metrics are calculated by comparing each candidate instruction to the three reference instructions given the navigation path. As suggested by~\cite{zhao2021evaluation}, SPICE is adopted as the primary metric.

\noindent\textbf{Evaluation Protocol.} As we implement our follower over different baselines (\ie, \textit{Seq2Seq}\!~\cite{anderson2018vision}, \textit{Speaker-Follower}\!~\cite{fried2018speaker}, and \textit{RCM}\!~\cite{fried2018speaker}), we report the performance of their corresponding speakers (\ie, Ours-\textit{Seq2Seq}, Ours-\textit{Speaker-Follower}, and Ours-\textit{RCM}). For comparative methods, we consider the speaker in~\cite{fried2018speaker} (\ie, BT Speaker) and the instruction generation model (\ie, VLS) in~\cite{agarwal2019visual}. The former is widely used in current VLN instruction following methods for data augmentation, and VLS is the only model that is specifically designed for instruction generation in VLN.

\noindent\textbf{Quantitative Result.} The comparison results on instruction generation are reported in Table~\ref{table:R2Rig}. As seen, our speakers achieve better performance on most of the metrics. For example, our three speakers gain the top three SPICE scores on both \texttt{val}~\texttt{seen} and \texttt{val}~\texttt{unseen} sets. Moreover, among our three speakers, Ours-\textit{RCM} performs best. We suspect that a stronger follower can provide better feedback signals to the speaker during cycle-consistent learning.

\noindent\textbf{User Study.} We organize two user studies. In the first user study, we sample 500 paths in \texttt{val} \texttt{unseen} and generate instructions through our three speakers (\ie, Ours-\textit{Seq2Seq}, Ours-\textit{Speaker-Follower}, Ours-\textit{RCM}). 25 volunteer students are asked to select the most meaningful instruction from those compared for each path. Ours-\textit{RCM} receives more votes than the others (Ours-\textit{RCM}: 40.5\% \textit{vs} Ours-\textit{Speaker-Follower}: 32.1\% \textit{vs} Ours-\textit{Seq2Seq}: 27.4\%). In the second user study, we let other 25 students compare the outputs of Ours-\textit{RCM}, BT Speaker and VLS. Ours-\textit{RCM} wins with 68.6\% picking rate (BT Speaker: 13.2\%, VLS: 18.2\%).
\vspace*{-2pt}
\subsection{Diagnostic Experiments}\label{sec:abl}
\vspace*{-3pt}
To fully examine the effectiveness of our core model designs, a set of ablative studies are conducted on \texttt{val} \texttt{unseen} of R2R \cite{anderson2018vision} (Table~\ref{table:abs}). Our model is built upon~\cite{fried2018speaker}.

\noindent\textbf{Cycle-Consistent Learning.} We first assess the contribution of our cycle-consistent learning scheme (\textit{cf.}~\S\ref{sec:2ag}). For the baseline method, the follower and speaker are trained separately and independently. Then we build four alternatives, \ie, $\triangle^A_E$, $\triangle^X_E$, $\triangle^A_{E\!}+_{\!}\triangle^X_E$, and $\triangle^A_E\!+\!\triangle^X_E\!+\!\triangle^{A'}_E$.  The first three baselines are trained by minimizing different $\triangle$ errors on the labeled dataset $\mathcal{D}$ only while the last one addi- tionally uses unlabeled data$_{\!}$~~$\mathcal{U}$, \ie, 17K randomly sampled\\
\noindent paths $A'$ as in~\cite{fried2018speaker}. From Table~\ref{table:abs}, we can conclude that: i) leveraging cross-task relations indeed boosts the performance on both instruction following and generation (while $\triangle^X_E$ greatly facilitates instruction generation with small performance sacrifice of instruction following), and ii) our cycle-consistency learning scheme not only works well on labeled data, but also makes better use of unlabeled data.

\noindent\textbf{Counterfactual$_{\!}$ Environment$_{\!}$ Creation.$_{\!}$} Next$_{\!}$ we$_{\!}$ study$_{\!}$~the efficacy$_{\!}$ of$_{\!}$ our$_{\!}$ counterfactual$_{\!}$ thinking$_{\!}$ strategy$_{\!}$ (\textit{cf.}$_{\!}$~\S\ref{sec:CEC}).$_{\!}$~To this end, we separately train the speaker and follower on~the counterfactual$_{\!}$ examples$_{\!}$ created$_{\!}$ by$_{\!}$ the$_{\!}$ creator$_{\!}$ without$_{\!}$ cycle-$_{\!}$ consistent learning. We build a baseline `\textit{w/o} reference environment $E^r$' by randomly masking a part of the original scene without considering reference environment. Table$_{\!}$~\ref{table:abs} shows that: i) our synthesized counterfactual environments can benefit the training of both the speaker and follower, and ii) additionally considering reference environments during synthesizing can produce more informative counterfactuals. 

\noindent\textbf{Full Model Design.} After examining the contribution of two critical components individually, we evaluate the effectiveness of our full model design. Our full model learns the speaker, follower, and creator jointly with comprehensive use of labeled and unlabeled ``real'' data as well as counterfactual examples. As evidenced in Table~\ref{table:abs}, CCC brings more significant performance improvements over each individual module on both instruction following and generation.

\noindent\textbf{Visual Comparison Results.} Finally, some visual results are provided in Fig.~\ref{fig:vr} for more intuitive comparisons. From the left sub-figure we can observe that, the follower trained with our CCC framework can derive a more robust navigation policy and thus reaches the target location successfully. As shown in the right sub-figure, our speaker is able to generate more accurate instructions; some novel actions and landmarks are successfully mentioned. During training, our strong speaker in turn boosts the learning of the follower.

\vspace*{-3pt}
\section{Conclusion}
\vspace*{-1pt}
We introduced a powerful training framework, CCC, that learns navigation generation and following simultaneously. It explicitly leverages cross-task connections to regularize the training of both the speaker and follower. Hence, a creator is integrated into such cycle-consistent learning system, so as to synthesize counterfactual environments and further facilitate training. The CCC framework is model-agnostic and can be integrated into a diverse collection of navigation models, leading to performance improvements over both instruction following and generation, in the R2R dataset.


\subfile{supp/VLN_supp-v3.tex}

{\small
\bibliographystyle{ieee_fullname}
\bibliography{xVLN}
}
\clearpage

\end{document}

%% file: supp/VLN_supp-v3.tex





\makesupptitle{Counterfactual Cycle-Consistent Learning for Instruction Following and Generation in Vision-Language Navigation \\\textit{Supplementary Material}}

This document provides more details of our approach and additional experimental results, organized as follows:
\begin{itemize}
	\vspace{-5pt}
	\setlength{\itemsep}{0pt}
	\setlength{\parsep}{0pt}
	\setlength{\parskip}{0pt}
	\item \S\ref{sec:I} \ Formula Derivation of Cycle-Consistent Training.
	\item \S\ref{sec:II} More Details of Counterfactual Cycle-Consistent Learning.
	\item \S\ref{sec:III} More Ablative Study on Counterfactual Environment Creator.
	\item \S\ref{sec:IV} Additional Qualitative Results.
	\item \S\ref{sec:discuss} Discussion about Social Impact and Limitations.
\end{itemize}

\section{Formula Derivation of Cycle-Consistent Training}\label{sec:I}
In this section, we introduce the derivation of gradient calculation for our cycle-consistent training in (\textit{cf.}~\S{\color{red}3.1}). Given aligned $(E, X)\!\in\!\mathcal{E}\!\times\!\mathcal{X}$, we can first obtain a navigation path $\hat{A}$ through the follower $f(E, X)$ (shortly $f_E(X)$). We then use the speaker to translate $\hat{A}$ to a visual-grounded instruction $s(E, \hat{A})$ (shortly $s_E(\hat{A})$), which is expected to be semantically similar to $X$, \ie, gaining a small  cycle-consistent error $\triangle_E(X, s_E(\hat{A}))$ (shortly $\triangle^X_E$). Similarly, for paired $(E, A)\!\in\!\mathcal{E}\!\times\!\mathcal{A}$, we have $\triangle^A_E$.  Then the $\triangle$ error can be specified as the negative log-likelihood, which is minimized for regularizing training (\ie, Eq.~{\color{red}6} in \S{\color{red}3.1}):
\vspace{-2pt}
\begin{equation}\small\label{eq:error_supp}
\begin{aligned}
	\!\!\!\!\!\!\triangle^A_E\!=\!&-\!\log\!\!\sum_{\hat{X}\in\mathcal{X}} \!\!P(s_E(A)\!=_{\!}\!\hat{X}|A)P(f_E(\hat{X})\!=_{\!}\!A|s_E(A)\!=_{\!}\!\hat{X}),\!\!\!\\
	\!\!\!\!\!\!\triangle^X_E\!=\!&-\!\log\!\!\sum_{\hat{A}\in\mathcal{A}} \!\!P(f_E(X)\!=_{\!}\!\hat{A}|X)P(s_E(\hat{A})\!=_{\!}\!X|f_E(X)\!=_{\!}\!\hat{A}).\!\!\!
\end{aligned}
\end{equation}
Eq.~\ref{eq:error_supp} can be briefly denoted as:
\begin{equation}\small
	\begin{aligned}
	\triangle^A_E=&-\log\sum\nolimits_{\hat{X}\in\mathcal{X}} P(\hat{X}|A; E)P(A|\hat{X}; E),\\
	\triangle^X_E=&-\log\sum\nolimits_{\hat{A}\in\mathcal{A}} P(\hat{A}|X; E)P(X|\hat{A}; E).
	\end{aligned}
\end{equation}
Directly calculating the gradients for $\triangle^A_E$ is intractable due to the huge space of $\mathcal{X}$. A similar issue also holds for  $\triangle^X_E$.
For ease of optimization, we optimize the upper bounds of $\triangle^X_E$ and $\triangle^A_E$, as in~\cite{wang2019multi}. According to Jensen's inequality, we have:
\begin{equation}
\begin{aligned}
	\bar{\triangle}^X_E=-\sum_{\hat{A}\in\mathcal{A}} P(\hat{A}|X;E)\log P(X|\hat{A};E) \geq \triangle^X_E,\\
	\bar{\triangle}^A_E=-\sum_{\hat{X}\in\mathcal{X}} P(\hat{X}|A;E)\log P(A|\hat{X};E) \geq \triangle^A_E.
\end{aligned}
\end{equation}

In this way, the gradient for $\bar{\triangle}^X_E$ and $\bar{\triangle}^A_E$ w.r.t. the parameters of speaker, \ie, $\theta^s$, can be derived as:
\begin{equation}
	\small
\begin{aligned}
	\frac{\partial\bar{\triangle}^X_E}{\partial\theta^s} = &-\sum_{\hat{A}\in\mathcal{A}} P(\hat{A}|X;E) \frac{\partial\log P(X|\hat{A};E)}{\partial \theta^s}\\
    = &-\mathbb{E}_{\hat{A}\sim P(\cdot|X;E)}\frac{\partial\log P(X|\hat{A};E)}{\partial \theta^s}\\
	\approx & - \frac{\partial\log P(X|\hat{A};E)}{\partial \theta^s},\\
	\frac{\partial\bar{\triangle}^A_E}{\partial\theta^s} =
	  &-\sum_{\hat{X}\in\mathcal{X}} \log P(A|\hat{X};E) \frac{\partial P(\hat{X}|A;E)}{\partial\theta^s}\\
	= &-\sum_{\hat{X}\in\mathcal{X}}\! P(\hat{X}|A;E)\! \log\! P(A|\hat{X};E) \frac{\partial\! \log\! P(\hat{X}|A;E)}{\partial\theta^s}\\
	= &-\mathbb{E}_{\hat{X}\sim P(\cdot|A;E)} [\log P(A|\hat{X};E) \frac{\partial \log P(\hat{X}|A;E)}{\partial\theta^s}]\\
	= &-\mathbb{E}_{\hat{X}\sim P(\cdot|A;E)} [(\log P(A|\hat{X};E)-b^f )\frac{\partial \log P(\hat{X}|A;E)}{\partial\theta^s}]\\
	\approx &- (\log P(A|\hat{X};E) - b^f) \frac{\partial \log P(\hat{X}|A;E)}{\partial\theta^s},
\end{aligned}
\end{equation}
where $b^f$ is a baseline to reduce the training variance and estimated by the mean of previous $\log P(A|\hat{X}; E)$~\cite{fu2019counterfactual}. Here $\log P(A|\hat{X};E)$ can also be regarded as a reward for the speaker to generate $\hat{X}$~\cite{he2016dual}.

\begin{table*}[t]
	\centering
\vspace*{-15pt}
			\resizebox{0.99\textwidth}{!}{
			\setlength\tabcolsep{6pt}
			\renewcommand\arraystretch{1}
	\begin{tabular}{c|c||cccc|cccccc}
	\hline 
	~ & & \multicolumn{4}{c|}{Instruction Following} & \multicolumn{6}{c}{Instruction Generation}\\
	\cline{3-12}
	\multirow{-2}{*}{} &\multirow{-2}{*}{Component} &\textbf{SR}$\uparrow$ &NE$\downarrow$ &OR$\uparrow$  &SPL$\uparrow$ &Bleu-1~$\uparrow$ &Bleu-4~$\uparrow$ &CIDEr~$\uparrow$ &Meteor~$\uparrow$ &Rouge~$\uparrow$ &\textbf{SPICE}~$\uparrow$ \\
	\hline
	\hline
	Baseline~\cite{fried2018speaker} & - & 29.9 &6.9  & 40.7 & 21.0 &0.522 &0.142 &0.114 &0.228 &0.346 &0.188\\
	\hline
	\multirow{3}{*}{\tabincell{c}{Training objective\\of creator $\mathcal{L}^c$}}  &$\mathcal{L}_{\ell_2}$ & 32.7 & 6.8 & 43.9 & 23.4 & 0.691 & 0.268 & 0.449 & 0.227 & 0.469 & 0.191\\
&$\mathcal{L}_{adv}$  & 38.3 & 6.4 & 47.1 & 27.3 & 0.703 & 0.272 & 0.469 & 0.228 & 0.470 & 0.198\\		\cline{2-12}
&$\mathcal{L}_{\ell_2}+\mathcal{L}_{adv}$ & \textbf{43.5} & \textbf{5.8} & \textbf{52.0} & \textbf{38.1} & \textbf{0.706} & \textbf{0.275} & \textbf{0.477} & \textbf{0.229} & \textbf{0.474} & \textbf{0.207}
	\\
	\hline
	\end{tabular}
	}
	\caption{The ablative study on the effect of each loss component of $\mathcal{L}^c$. The baseline agent is \textit{speaker-follower}~\cite{fried2018speaker}.}
	\label{tab:abl_supp}
\end{table*}

In the same way, the gradient for $\bar{\triangle}^A_E$ and $\bar{\triangle}^X_E$ w.r.t. the parameters of follower, \ie, $\theta_f$, are given as:
\begin{equation}
	\small
\begin{aligned}
	\frac{\partial\bar{\triangle}^A_E}{\partial\theta^f} \approx & - \frac{\partial\log P(A|\hat{X};E)}{\partial \theta^f},\\
	\frac{\partial\bar{\triangle}^X_E}{\partial\theta^f} \approx &- (\log P(X|\hat{A};E) - b^s) \frac{\partial \log P(\hat{A}|X;E)}{\partial\theta^f},
\end{aligned}
\end{equation}
where $b^s$ is a baseline to reduce the training variance and estimated by the mean of previous $\log P(X|\hat{A};E)$.

\section{Details of Counterfactual Cycle-Consistent Learning}
\label{sec:II}
To illustrate the details of Counterfactual Cycle-Consistent (CCC) learning, we present the pseudo code of CCC learning on labeled data in Algorithm \ref{alg:ccc}. For CCC learning on unlabeled data, we can modify the algorithm according to Eq.~\textcolor{red}{9} in \S\textcolor{red}{3.1}.

\begin{algorithm}
    \caption{The pseudo code of Counterfactual Cycle-Consistent (CCC) Learning on labeled data.}\label{alg:ccc}
    \textbf{Arguments:} The labeled dataset $\mathcal{D}\!=\!\{(E,X,A)\}$, the maximum iteration $N$, the learning rate $\alpha$, the parameters of speaker $\theta^s$, follower $\theta^f$ and creator $\theta^c$.
    \begin{algorithmic}[1]
    \State Initialize the parameters $\theta^s,\theta^f,\theta^c$
    \For{iteration $i \in [1,\dots,N]$}
		\State Sample batch $B\subset \mathcal{D}$
		\State $\theta^{s}_{\text{tmp}}\gets \theta^s$
		\State $\theta^{f}_{\text{tmp}}\gets \theta^f$
		\State $\theta^{c}_{\text{tmp}}\gets \theta^c$
        \For{$(E,X,A)\in B$}
			\State $\hat{X}\sim P_{\theta^s}(\cdot|A)$
			\State $\hat{A}\sim P_{\theta^f}(\cdot|X)$
			\State Estimate $\triangle_E^A$ and $\triangle_E^X$ \Comment{Defined in Eq.\textcolor{red}{6}.}
			\State $\theta^{s}_{\text{tmp}}\gets \theta^{s}_{\text{tmp}}-\frac{\alpha\partial(\triangle_E^A+\triangle_E^X)}{\partial \theta^s}$
			\State $\theta^{f}_{\text{tmp}}\gets \theta^{f}_{\text{tmp}}-\frac{\alpha\partial(\triangle_E^A+\triangle_E^X)}{\partial \theta^f}$
			\State Create $\bar{E}$ \Comment{See Eq.\textcolor{red}{11}.}
			\State Estimate $\mathcal{L}^c$ \Comment{Defined in Eq.\textcolor{red}{12}.}
			\State $\theta^{c}_{\text{tmp}}\gets \theta^{c}_{\text{tmp}}-\alpha\frac{\partial\mathcal{L}^c}{\partial \theta^c}$

		\EndFor

        \State $\theta^{s}\gets \theta^s_{\text{tmp}}$
		\State $\theta^{f}\gets \theta^f_{\text{tmp}}$
		\State $\theta^{c}\gets \theta^c_{\text{tmp}}$
    \EndFor
    \State \textbf{return} $\theta^s,\theta^f,\theta^c$
    \end{algorithmic}
\end{algorithm}

\section{More Ablative Study on Counterfactual Environment Creator}\label{sec:III}
To further reveal the effectiveness of each component of $\mathcal{L}^c$, we train the agent \textit{w} and \textit{w/o} $\mathcal{L}_{l_2}$ and $\mathcal{L}_{adv}$ term. As shown in Tab.~\ref{tab:abl_supp}, compared to the baseline agent, the agent equipped with the counterfactual environment creator achieves better performance on both instruction following and instruction generation task. Specifically, in the second line, without the adversarial term $\mathcal{L}_{adv}$, the performance of the agent drops compared to the full model because the creator fails to preserve the feasibility of the created environment. In the third line, the agent is trained without the $l_2$-norm loss that constrains the similarity between the original environment and the created environment. The performance of the agent finally suffers from the limited variation of the created environment.

\section{Additional Visual Comparison Results}
\label{sec:IV}
\begin{figure*}[!tbh]
	  \centering
		  \includegraphics[width=0.85\linewidth]{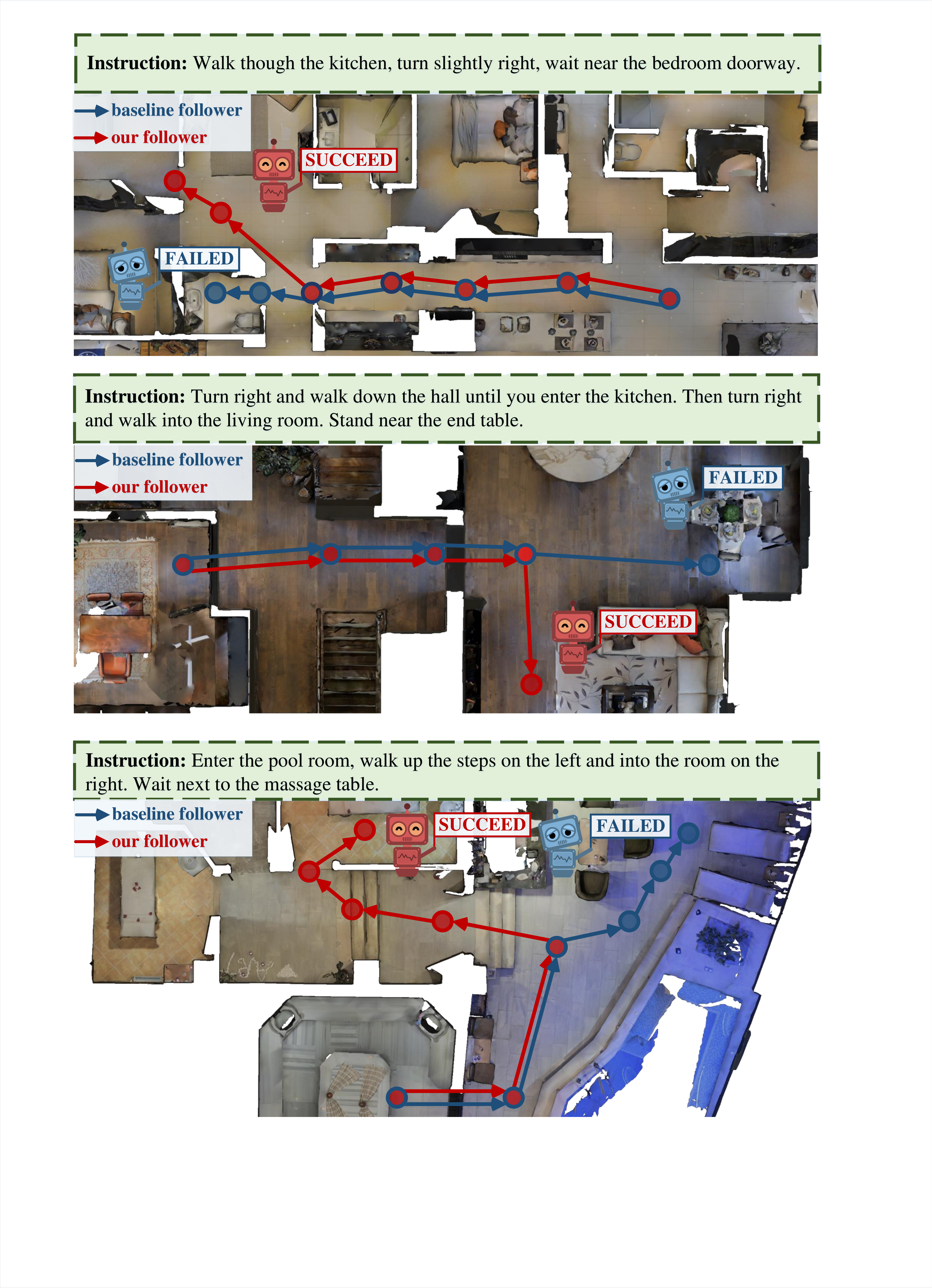}
	\captionsetup{font=small}
		\caption{\small{Visual comparison results on instruction following between the baseline follower and our follower.}
	}
\label{fig:vr_supp}
\end{figure*}

\begin{figure*}[!tbh]
	  \centering
		  \includegraphics[width=0.91\linewidth]{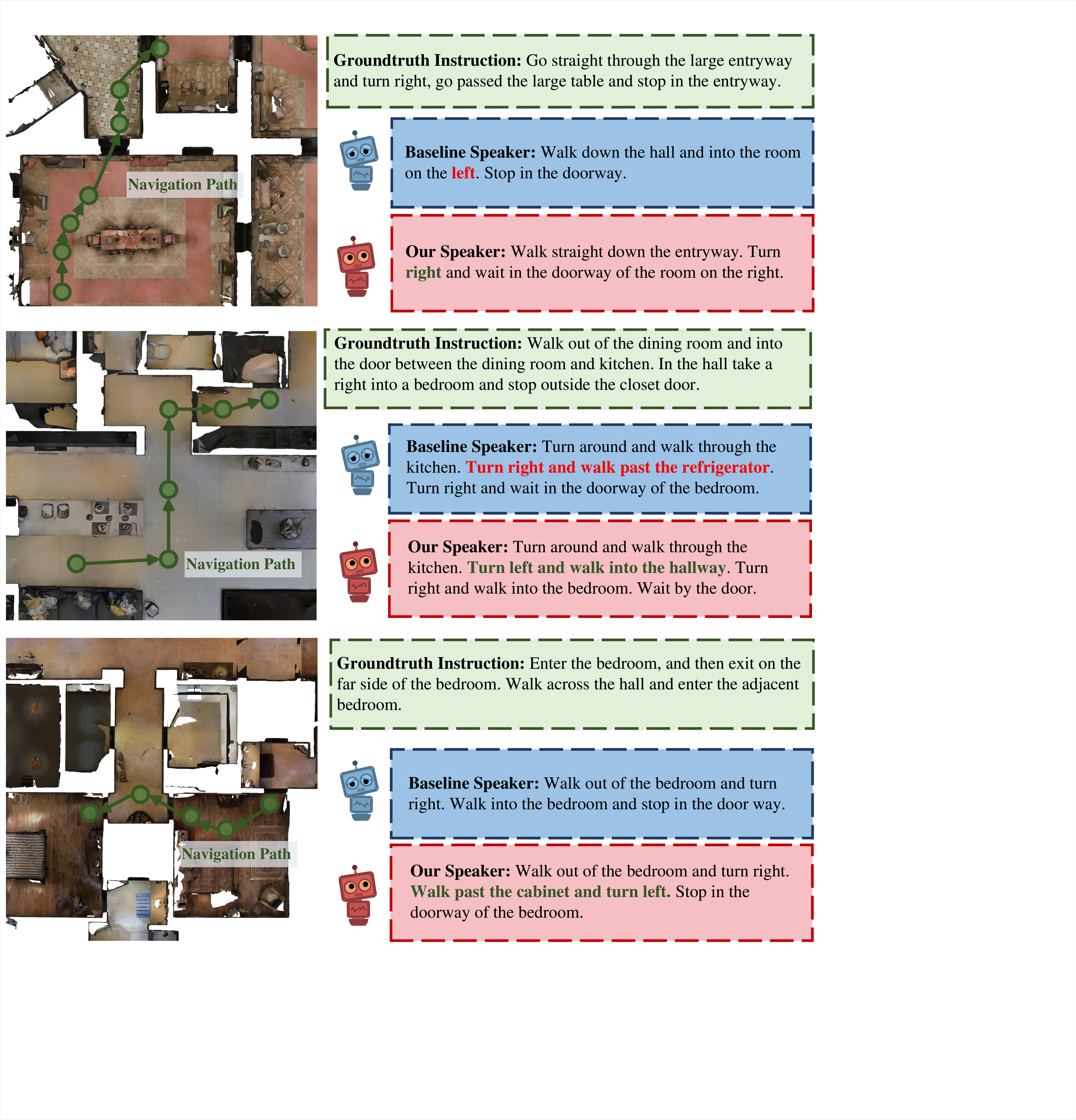}
	\vspace{-5pt}
	\captionsetup{font=small}
		\caption{\small{Visual comparison results on instruction  generation. The inconsistent descriptions generated by the baseline speaker are high-lighted in \textcolor{red}{\textbf{red}}. The corresponding crucial accurate descriptions generated by our speaker are high-lighted in \textcolor{dgreen}{\textbf{dark green}}.
	}
	}
\label{fig:ig_supp}
\vspace{-14pt}
\end{figure*}
In this section, we provide more qualitative results for instruction following and generation. The speaker and follower are implemented as~\cite{fried2018speaker}, while the baseline agents are trained separately following~\cite{fried2018speaker} and ours are trained with our CCC framework.

\noindent \textbf{Instruction Following.} The visualization of navigation trajectories is shown in Fig.~\ref{fig:vr_supp}. We observe that the follower trained with our CCC framework learns a robust policy and successfully reaches the target locations by following the given instructions.

\noindent \textbf{Instruction Generation.} The generated instructions are illustrated in Fig.~\ref{fig:ig_supp}. We find that the speaker trained with our CCC framework generates more accurate and detailed instructions. Concretely, compared to the baseline speaker, our CCC speaker is able to (1) comprehensively describe the whole action sequence, (2) highlight crucial landmarks, and (3) mention more essential details.

\section{Discussion}
\label{sec:discuss}
\noindent\textbf{Social Impact.}
Current agents are developed in virtual static environments. If the algorithm is deployed on a real robot in a real  dynamic environment, the collisions during navigation can potentially cause damage to persons and assets. To prevent risky collisions, the robot needs to move under some security restrictions.

\noindent\textbf{Limitations.} Our work focuses on interior Vision-and-Language Navigation task. The generalization of this approach to other navigation tasks is not clear. We would verify it in future work.

